\documentclass[conference]{IEEEtran}
\usepackage{times}

\usepackage[numbers]{natbib}
\usepackage{multicol}
\usepackage[bookmarks=true]{hyperref}

\ifCLASSINFOpdf
  \usepackage[pdftex]{graphicx}
\else
  \usepackage[dvips]{graphicx}
\fi
\usepackage{placeins}
\usepackage{booktabs}
\usepackage{multirow}

\usepackage{amsmath}
\usepackage{amssymb}  
\usepackage{mathtools}
\usepackage{siunitx}
\usepackage{bbm}
\usepackage{bm}

\newcommand{\vect}{\bm}	

\usepackage{xcolor}

\begin{document}

\title{Design and Control of a Bipedal Robotic Character} 

\author{\authorblockN{Ruben Grandia\authorrefmark{1},
Espen Knoop\authorrefmark{1},
Michael A. Hopkins\authorrefmark{2},
Georg Wiedebach\authorrefmark{2},\\
Jared Bishop\authorrefmark{3},
Steven Pickles\authorrefmark{3},
David M\"uller\authorrefmark{1}, and
Moritz B\"acher\authorrefmark{1}}
\authorblockA{\authorrefmark{1}Disney Research, Switzerland,
\authorrefmark{2}Disney Research, USA,
\authorrefmark{3}Walt Disney Imagineering R\&D, USA}}

\maketitle

\begin{abstract}
Legged robots have achieved impressive feats in dynamic locomotion in challenging unstructured terrain. 
However, in entertainment applications, the design and control of these robots face additional challenges in appealing to human audiences.
This work aims to unify expressive, artist-directed motions and robust dynamic mobility for legged robots. 
To this end, we introduce a new bipedal robot, designed with a focus on character-driven mechanical features. 
We present a reinforcement learning-based control architecture to robustly execute artistic motions conditioned on command signals.
During runtime, these command signals are generated by an animation engine which composes and blends between multiple animation sources.
Finally, an intuitive operator interface enables real-time show performances with the robot. 
The complete system results in a believable robotic character, and paves the way for enhanced human-robot engagement in various contexts, in entertainment robotics and beyond.
\end{abstract}

\IEEEpeerreviewmaketitle

\FloatBarrier

\section{Introduction}

\begin{figure}[t]
    \centering
    \includegraphics[trim={50 0 0 0},clip,width=\linewidth]{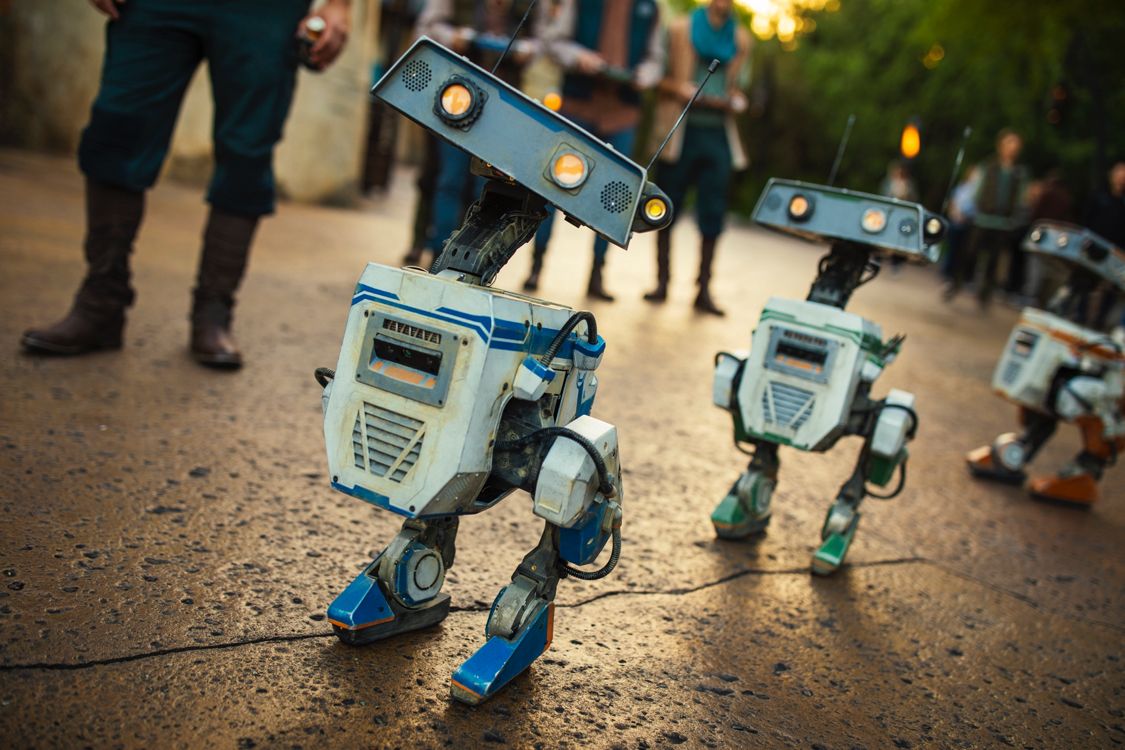}
    \caption{Three instances of our robotic character performing an unscripted show. Apart from their theming, they are identical. Each robot is remote-controlled by a separate operator.}
    \label{fig:teaser}
\end{figure}

Legged robotic platforms have gained widespread accessibility and are often envisioned as versatile mobile platforms suitable for navigating challenging and unstructured environments. Consequently, most robotic systems are engineered and controlled with utility and efficiency as the primary objectives~\cite{hurst2019walk}. This has led to remarkable achievements in dynamic legged systems that can now hike up mountains~\cite{miki2022learning} and conquer obstacle courses~\cite{hoeller2023anymal}.

Nevertheless, robots are increasingly being deployed beyond isolated environments, in places where they engage directly with humans. A rich set of applications can be found in collaborative robots~\cite{villani2018survey}, companion robots~\cite{robinson2013psychosocial}, art~\cite{kroos2012evoking}, and entertainment~\cite{fujita2001aibo}. These applications introduce additional challenges to the design and control of robots, as the success of the robot is additionally dependent on the subjective perception of humans. Legged systems must perform these expressive motions while simultaneously satisfying the already complex kinematic and dynamic balancing requirements inherent to whole-body locomotion.

Animators have mastered the art of breathing life into digital characters \cite{multon1999computer}, and a long-standing aspiration has been to endow robots with an equivalent level of expressiveness~\cite{van2004bringing}. Learning-based approaches to physics-based characters have made it possible to learn control policies from animation or motion capture input~\cite{peng2018deepmimic,peng2020learning}. Still, significant effort is required to translate such a performance from simulation to the real world. Meanwhile, other studies explore how robots can spark an emotional response through facial expressions, visual cues, or body language, although many of the robots used have limited mobile capabilities~\cite{venture2019robot}. 

In this work, we aim to bring expressive and dynamic motions onto a bipedal robotic character, and explore the intersection of legged robot design, control, and character animation. We present a new robot character, shown in Fig~\ref{fig:teaser}, with a mechanical design that is primarily driven by creative intent and simplicity rather than functional requirements. Additionally, we present a complete pipeline centered around reinforcement learning to bring animations onto the physical system. Our pipeline allows the robot to imitate artistic motions, and, based on user input, blend or seamlessly transition between them while remaining robust to uncertainty and external disturbances. Finally, we propose an intuitive \textit{puppeteering} interface that combines direct motion control with expressive animations. Through this interface, an operator is able to author believable interactions in real-time. The complete system allows us to rapidly explore robotic performances in an entertainment context. Similar building blocks could be used to create more expressive autonomous robots.

Succinctly, in addition to the presented system as a whole, contributions of this work include:
\begin{itemize}
    \item A workflow that integrates animation content, design, control, and real-time puppeteering, and enables the rapid development of custom robot characters.\footnote{The presented robot was developed by the authors in less than a year}.
    \item A new robot whose morphology and kinematics are driven by creative intent, rather than by function.
\end{itemize} 

\begin{figure*}[ht!]
  \includegraphics[trim={45 520 70 60},clip,width=\textwidth]{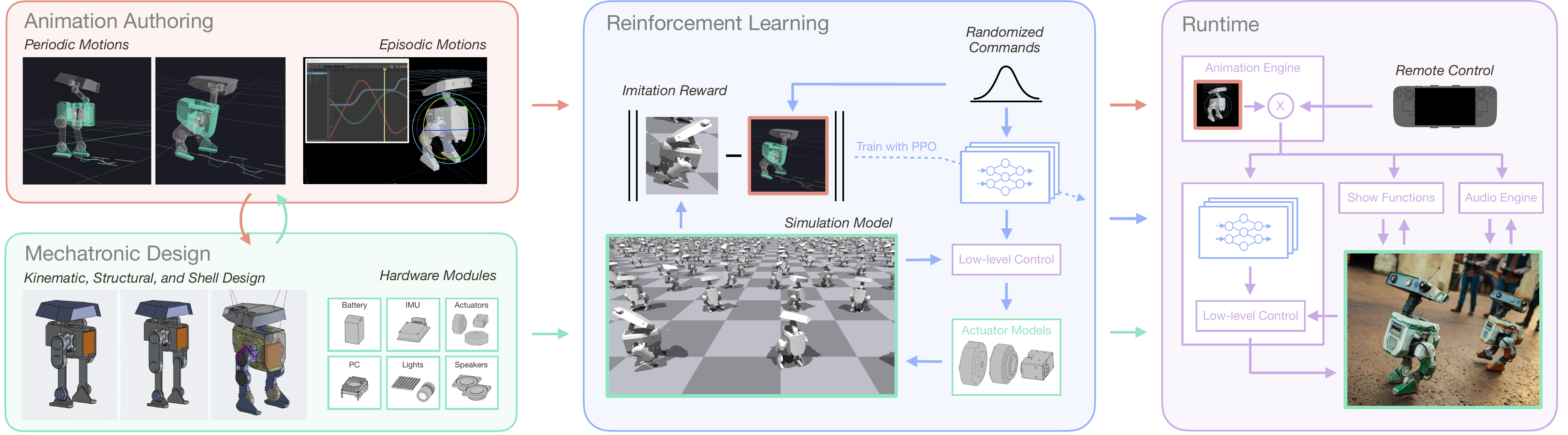}
  \caption{Our character design and control pipeline consists of animation, mechatronic design, reinforcement learning, and run-time tools. Animation and mechatronic design form an iterative process to define the character and its motion repertoire. These inputs are used in reinforcement learning which, through imitation rewards, results in control policies that robustly execute the intended motions, conditioned on external commands. During run-time, the animation engine combines user-inputs and animation content and interfaces with the control policies. In parallel, the animation engine synchronizes show-functions with the motion. }
  \label{fig:overview}
\end{figure*}

\begin{itemize}
    \item A breakdown of different motions into separate categories and control policies, switched between at runtime.
    \item A puppeteering interface that leverages conditional policy inputs, and enables robot performances by layering, blending and switching between different motion elements. 
\end{itemize}

\section{Related Work}

Several robots, initially intended for high-performance dynamic locomotion, have been used to perform expressive motions~\cite{li2020animated}, particularly in the context of dancing~\cite{bi2018realtime,bostondynamics2021how}. On the other hand, various robots have been purposefully designed for social interaction with humans~\cite{ishiguro2001robovie,kkedzierski2013emys}. Notably, humanoids such as iCub~\cite{metta2010icub}, NAO~\cite{gouaillier2009mechatronic}, and Pepper~\cite{pandley2019massproduced}, have been widely used in the research community, serving both as platforms for motion control and human-robot interaction~(HRI) research. 

However, these robots in previous studies were conceived as general-purpose platforms, and their demonstrations were often conducted as part of distinct projects or by entirely different organizations. Furthermore, their humanoid form naturally drives the mechanical design and comes with abundant human reference motions to be used for control. In this work, we introduce a robot where both the mechanical design and the motion that it performs are co-developed and driven by an artistic vision to create a unified character.

\subsection{Animating Robots}

The influence of a robot's movements on human perception has been widely studied and acknowledged~\cite{van2014robot}.
Van Breemen proposed to apply the principles of animation to bring robots to life~\cite{van2004bringing}, and presented an \textit{animation engine} that executes, composes, and blends multiple animations based on external commands~\cite{van2004animation}. Similarly, Fujita \textit{et al.} outlined a software architecture for entertainment robots~\cite{fujita1997open}, applied to robots like the quadrupedal {AIBO}~\cite{fujita2001aibo} and the humanoid {SDR-4X}~\cite{fujita2003autonomous}. Later, for the {NAO} platform, a graphical tool was introduced to make behavior programming more accessible~\cite{pot2009choregraphe}. In this paper, we leverage the motion composition and blending principles outlined in these works. However, we do not present a fully autonomous system that plans and executes long-term behaviors. Instead, we leave such high-level decisions to a puppeteer. With the proposed puppeteering interface and control policies that accept high-level inputs, we strive for a balance between the imitation of fixed artist-created motion and real-time interactive show authoring. We compose continuous inputs like the robot gaze, triggered animations, and autonomous stylized motion that is embedded in our low-level controller.

\subsection{Controlling Legged Systems}

A key challenge in controlling legged robots are the complex requirements related to the kinematics and under-actuated dynamics of the system. To navigate the constrained space of feasible solutions, model-based optimization approaches have been proposed as a tool to convert animations into dynamically feasible reference trajectories~\cite{suleiman2008human,nakaoka2010intuitive,li2020animated,grandia2023doc}. Still, an online controller is needed to stabilize the system as it inevitably deviates from the planned trajectory, for example through Model Predictive Control (MPC)~\cite{bjelonic2022offline}. We use model-based tools during the design phase to study the performance of the character and interpolate animated walk cycles.

Alternatively, Reinforcement Learning (RL) has become a popular choice for synthesizing closed-loop control policies directly from imitating reference motions~\cite{peng2018deepmimic,peng2020learning,hasenclever2020comic}. In a similar spirit, RL has been used to imitate solutions from a model-based motion planner~\cite{kang2023rl} or gait library~\cite{li2021reinforcement}. Clustering motions and learning a mixture-of-experts have been proposed to scale these methods to a large and diverse set of motions. Alternatively, an adversarial reward can be used instead of tracking rewards to imitate behaviors from unstructured large scale human motion datasets~\cite{peng2021amp,escontrela2022adversarial}. However, for our unique character such a dataset is not readily available.

While most related work strives for a single control policy, we adopt a divide-and-conquer strategy and train separate policies to imitate artist-authored walking, standing, and short animation sequences, conditioning the individual policies on carefully chosen high-level control commands. When paired with a real-time compositing and blending layer, these commands then enable the seamless and intuitive puppeteering of the robot with a two-joystick remote controller.

\section{Overview}

Our character design and control workflow, outlined in Fig.~\ref{fig:overview}, starts with an iterative process between mechanical design and animation. Classical animation tools, which use a rig consisting of links and spherical joints, are used to study the character and view it in the context of a scene. These studies inform the general proportions and range of motion of the character. A procedural gait generation tool is used to create periodic walking cycles. This tool builds on the rigid body dynamics of the system, and therefore generates physically plausible motions. The joint positions, velocities, and torques feed back into the mechanical design to refine geometry, select actuators, and perform structural analysis. This combination of tools allows us to rapidly explore motions, walking styles, and mechanical designs, to ultimately find the right trade-off between the physical limits of available hardware modules and the creative intent.

We strive not for the best possible mechanical design, but for a simple design which satisfies the creative intent. By using off-the-shelf hardware modules and characterizing them well in software, we can reduce hardware complexity while maintaining system performance.

After converging on an initial set of motions and mechanical design, both are used to define a reinforcement learning problem. The mechanical design forms the simulation model, together with actuator models and domain randomization. Kinematic motion references are exported from the animation tools and used in imitation rewards that maximize the similarity between simulated and reference motion. A set of commands are defined that provide high-level control of the robot. We then train multiple policies, one per motion or motion type, and condition them on these commands. This process results in policies that can robustly execute the defined animations while providing control over the character through policy switches and command signals.

During run-time, the animation engine receives user-input from a remote control interface. The animation engine fuses these inputs with predefined animations to generate commands for the control policies. Additionally, the animation engine triggers policy switches. Show functions and audio are controlled elements of the robot that play a key role in expressing character, but do not affect the dynamics of the system. Their behavior is synchronized with the motion of the robot through animation signals from the animation engine and state feedback.

\section{Mechatronic Design}

\begin{figure}[tb]
    \centering
    \includegraphics[width=\linewidth]{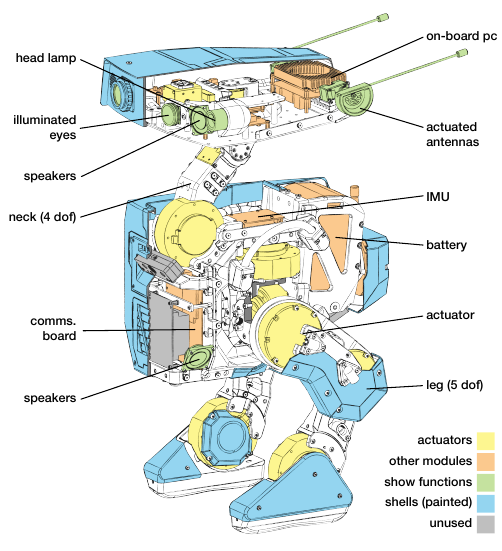}
    \caption{Mechanical design of our robotic character. The robot has 5 degrees of freedom per leg. The neck and head assembly has 4 degrees of freedom. The torso contains a custom communication board, a battery module, and an IMU. The onboard PC, radio receiver, and show function board are located in the head. Show functions consists of antennas, LED arrays as eyes, and a head lamp. A pair of speakers is located in the head and another pair in the bottom of the torso.}
    \label{fig:mechanics_annotated}
\end{figure}

Our final bipedal robotic character design is shown in Fig.~\ref{fig:mechanics_annotated}. The robot has 5 degrees of freedom (DoF) per leg and a 4 DoF neck and head assembly. The large workspace of the two legs enables a wide range of dynamic locomotion and lower body motions while the head can be posed independently relative to the torso. Similar to ANYmal~\cite{hutter2016anymal}, we adopted a design where actuators are directly placed at joints. To build our custom robot quickly, we 3D printed the components that connect pairs of off-the-shelf actuators. As a result of this design philosophy, the robot has the ankle actuators directly placed at the two feet and is not equipped with ankle roll actuators. To allow for passive ankle roll, we rounded off the two foot soles. By molding them with urethane foam, foot impacts on hard ground are further dampened. The knee joints bend backwards as creatively envisioned.

The robot has a total mass of \SI{15.4}{\kilo\gram}, with the torso weighing \SI{5.8}{\kilo\gram}, the neck and head \SI{2.4}{\kilo\gram}, and each leg \SI{3.6}{\kilo\gram}. It is \SI{0.66}{\meter} tall (excluding antennas) and the legs have a nominal length of \SI{0.28}{\meter} and an extended length of \SI{0.34}{\meter}.

The hip-adduction-abduction, hip-flexion-extension, and knee actuators are strongest with a peak torque of \SI{34}{\newton\meter} and maximum velocity of \SI{20}{\radian\per\second}. The hip-rotation, ankle, and the lower neck actuator have a peak torque of \SI{24}{\newton\meter} and maximum velocity of \SI{30}{\radian\per\second}. These two types are quasi-direct drive actuators and support high-bandwidth open-loop torque control suitable for dynamic locomotion~\cite{wensing2017proprioceptive}. The 3 actuators located in the head have a high gear ratio with a peak torque of \SI{4.8}{\newton\meter} and maximum velocity of \SI{6.3}{\radian\per\second}.

A custom microcontroller-driven communications board serves as the interface between the on-board PC, actuators, and an Inertial Measurement Unit (IMU), providing a communication rate of \SI{600}{\hertz}. All actuators have integrated drives which implement a low-level control loop and report motor position as measured by their built-in encoders. The on-board PC communicates with the hand-held operator controller through redundant wireless communication using both WiFi and LoRa radios. A removable battery powers the robot for at least \SI{1}{\hour} of continuous operation.

What sets our robot apart from other legged systems is a set of show functions: a pair of actuated antennas and illuminated eyes, and a head lamp. These functions provide animators with an additional means to express emotions. However, since they do not affect the system dynamics and are controlled in an open-loop fashion, we treat them separately from the main actuators. This allowed us to easily add and remove show functions during the design phase of the robot. In addition to these functions, we equipped the robot with a stereo pair of loudspeakers in both the body and the head.

\begin{figure*}[tb]
{
\footnotesize
\setlength{\tabcolsep}{1pt}
\begin{tabular}{ccccc}
\includegraphics[trim={550 100 550 500},clip,width=0.23\linewidth]{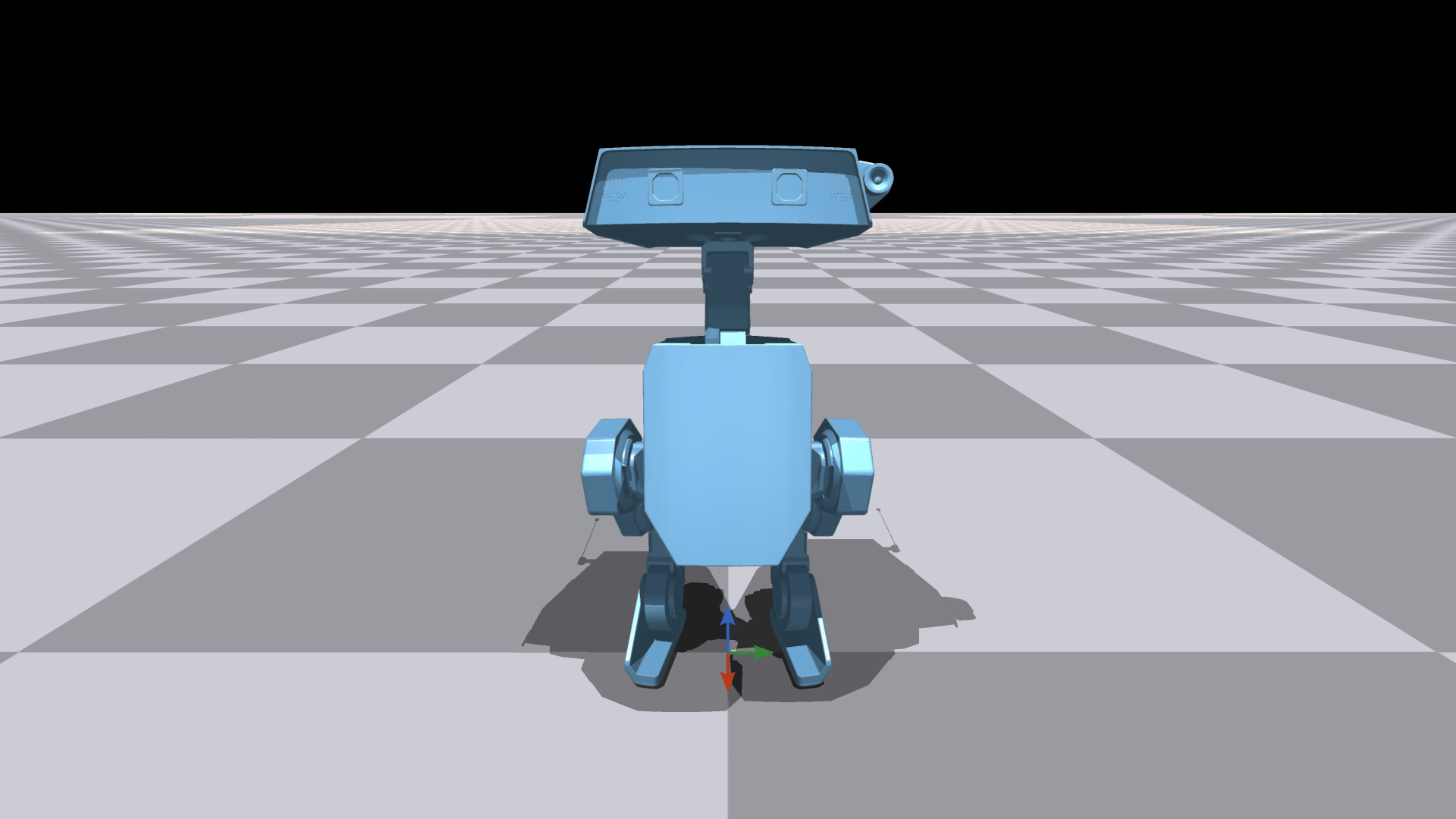} &
\includegraphics[trim={250 0 250 0},clip,width=0.187\linewidth]{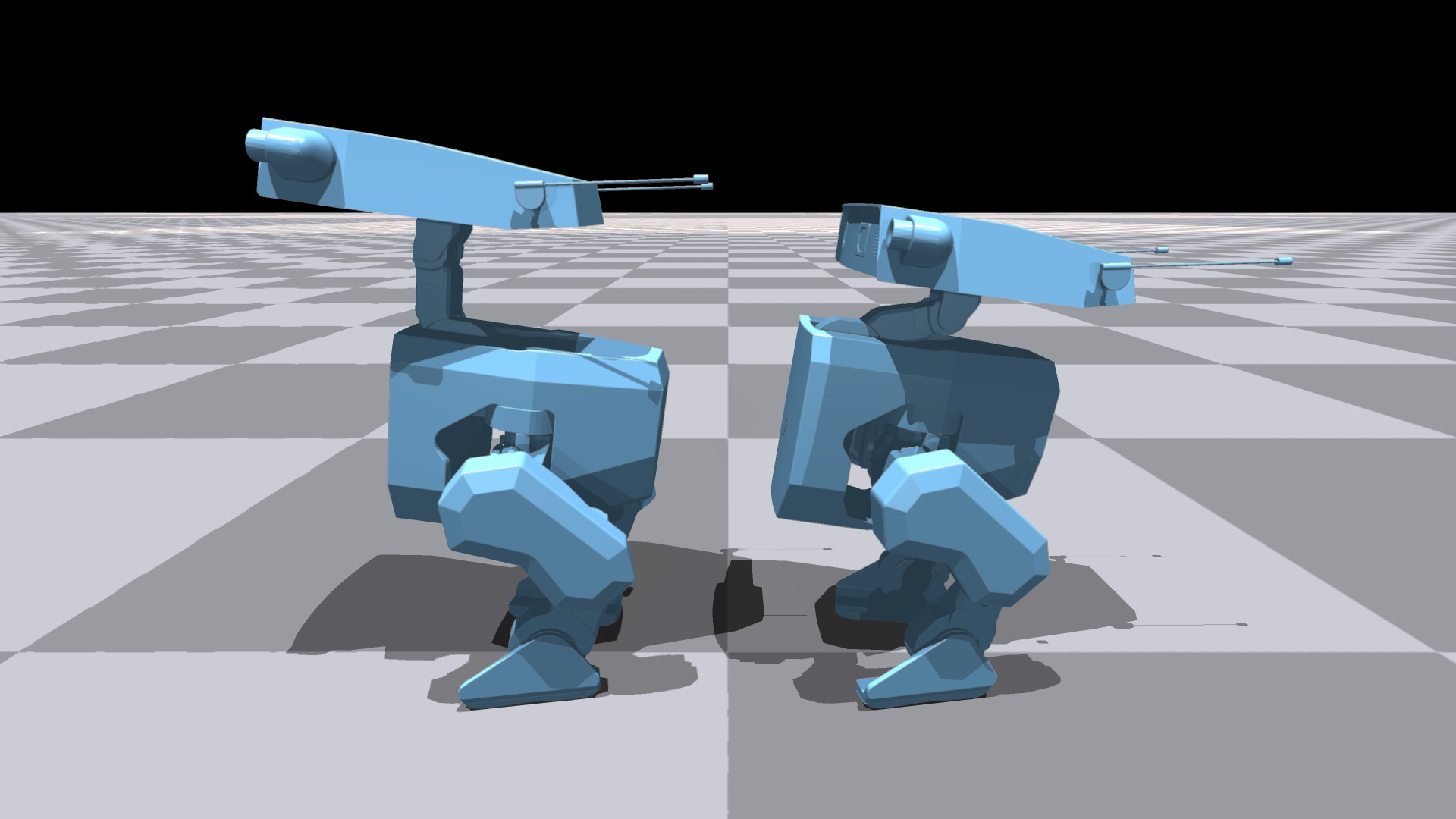} & 
\includegraphics[trim={250 0 250 0},clip,width=0.187\linewidth]{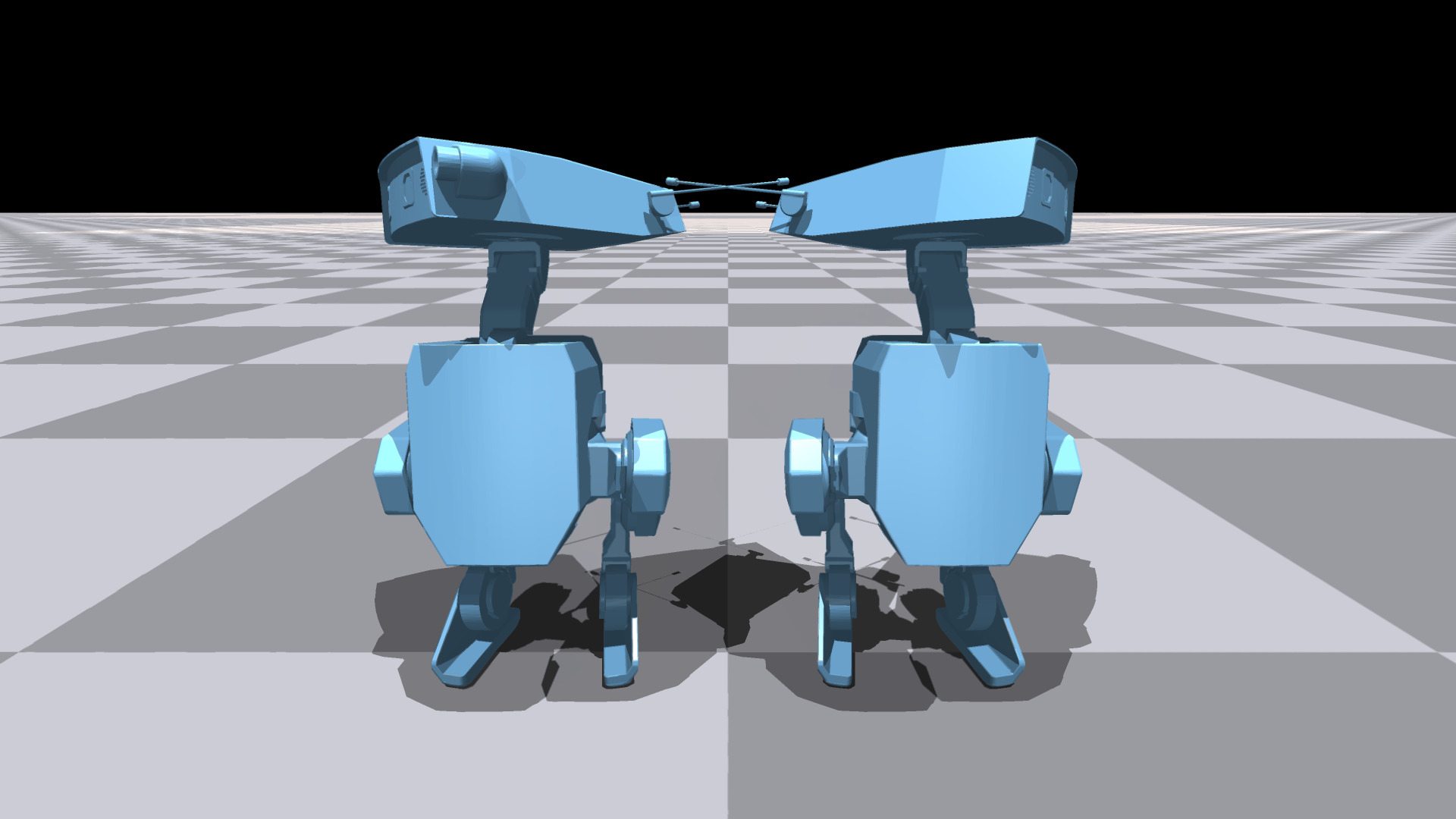} & 
\includegraphics[trim={250 0 250 0},clip,width=0.187\linewidth]{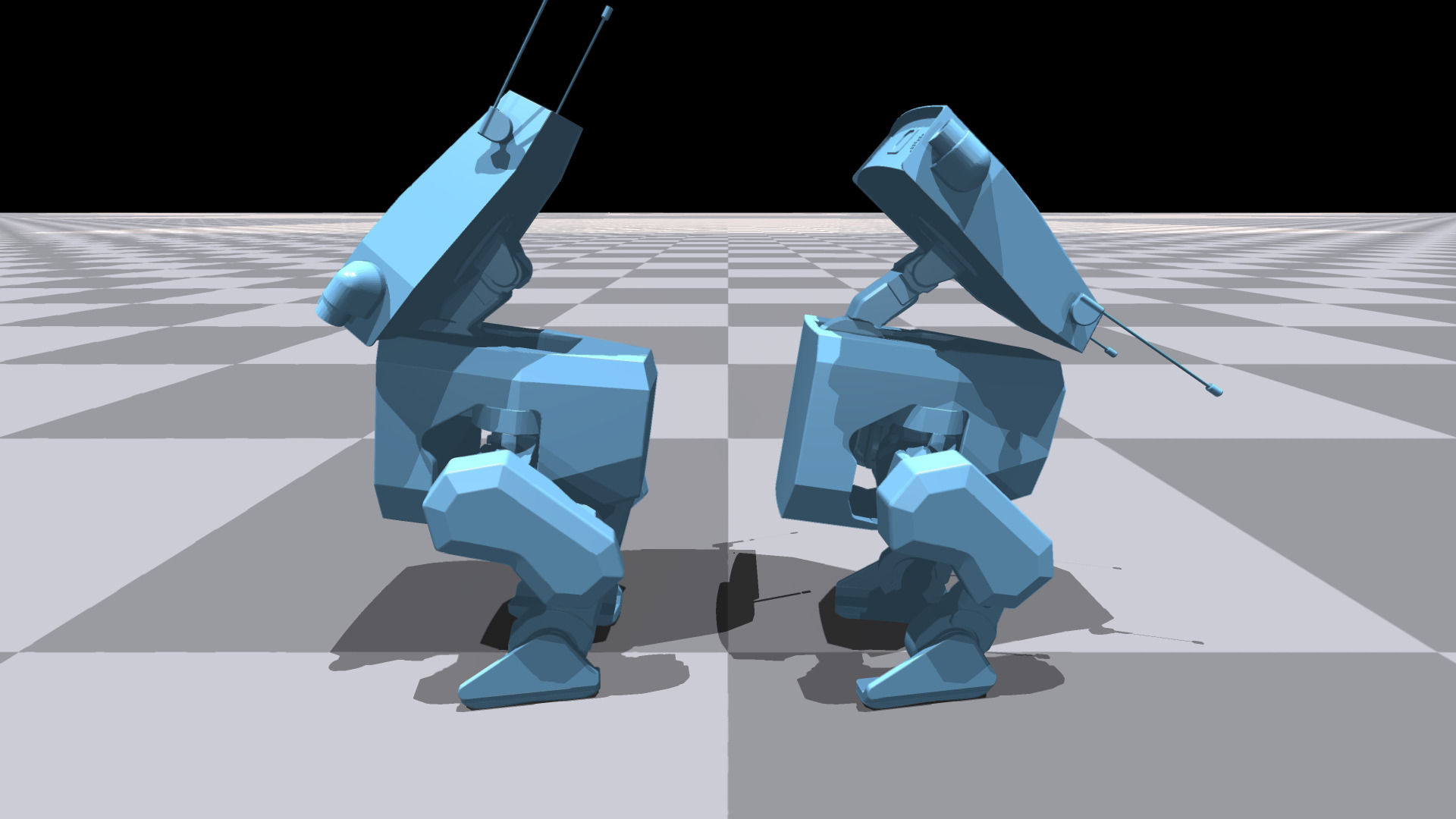} & 
\includegraphics[trim={250 0 250 0},clip,width=0.187\linewidth]{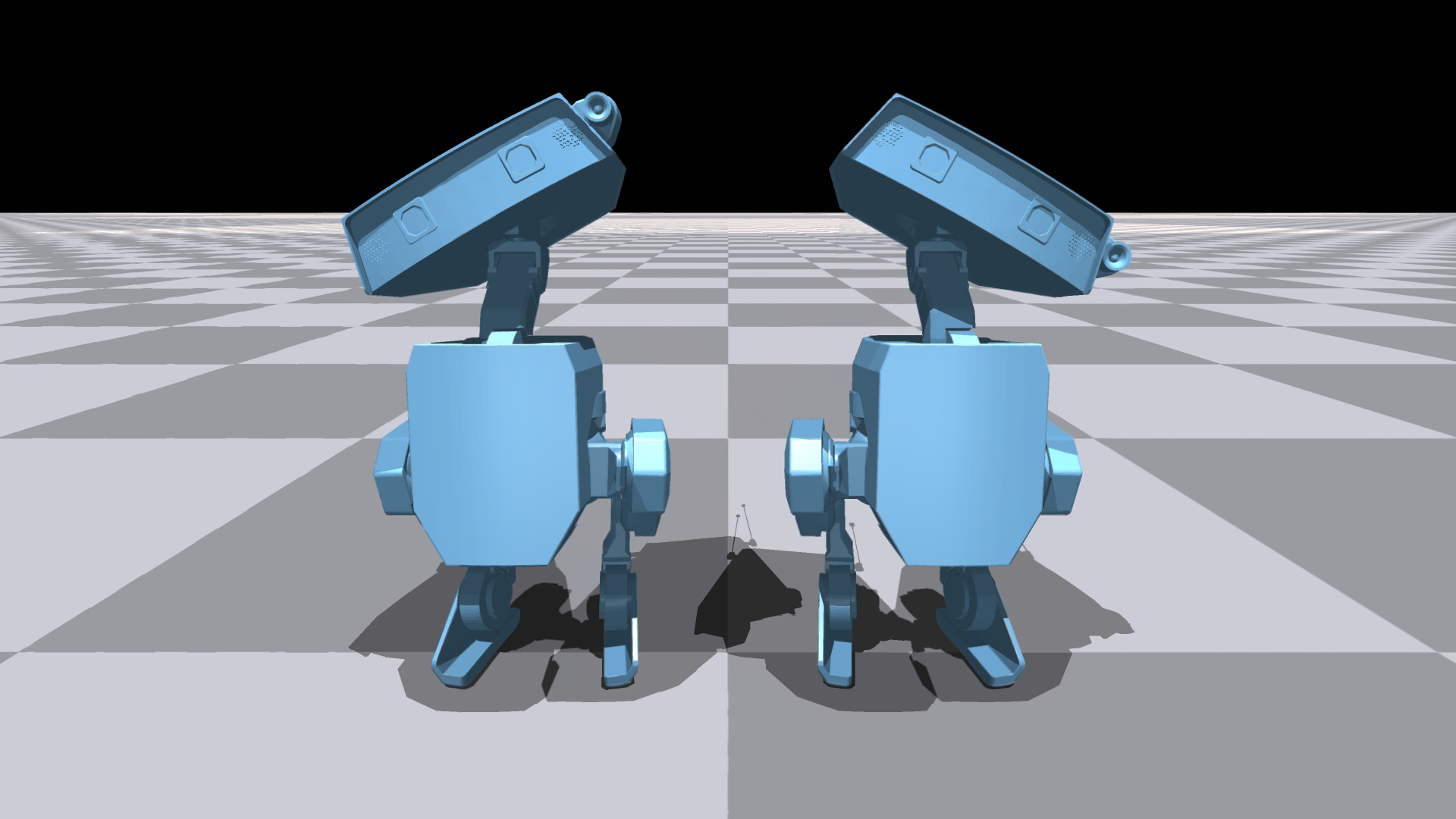} \\ 
\includegraphics[width=0.23\linewidth]{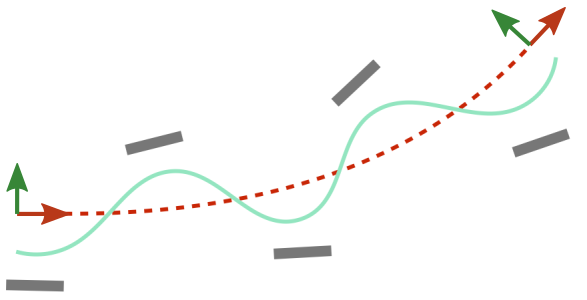} &
\includegraphics[trim={250 0 250 0},clip,width=0.187\linewidth]{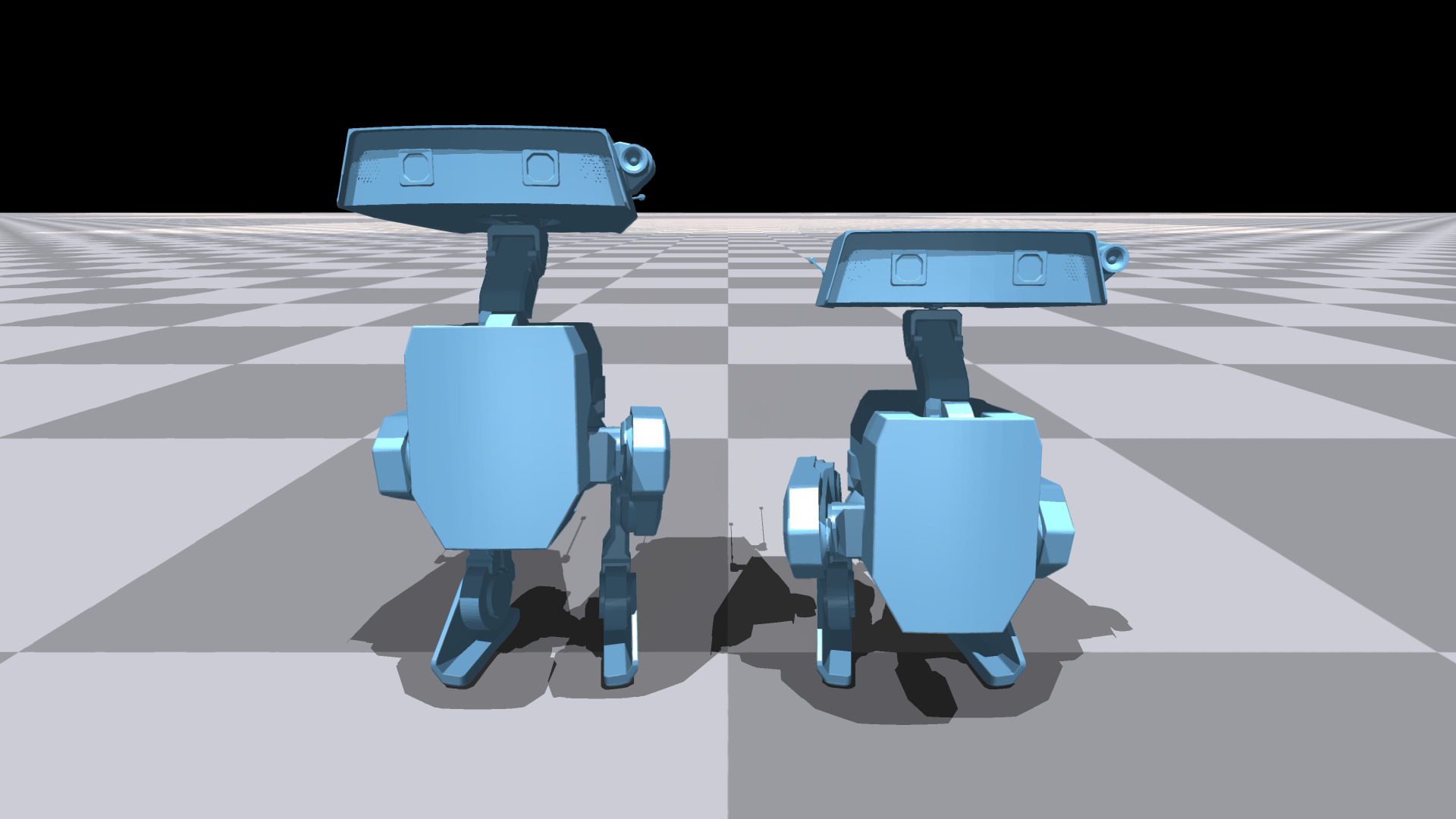} & 
\includegraphics[trim={250 0 250 0},clip,width=0.187\linewidth]{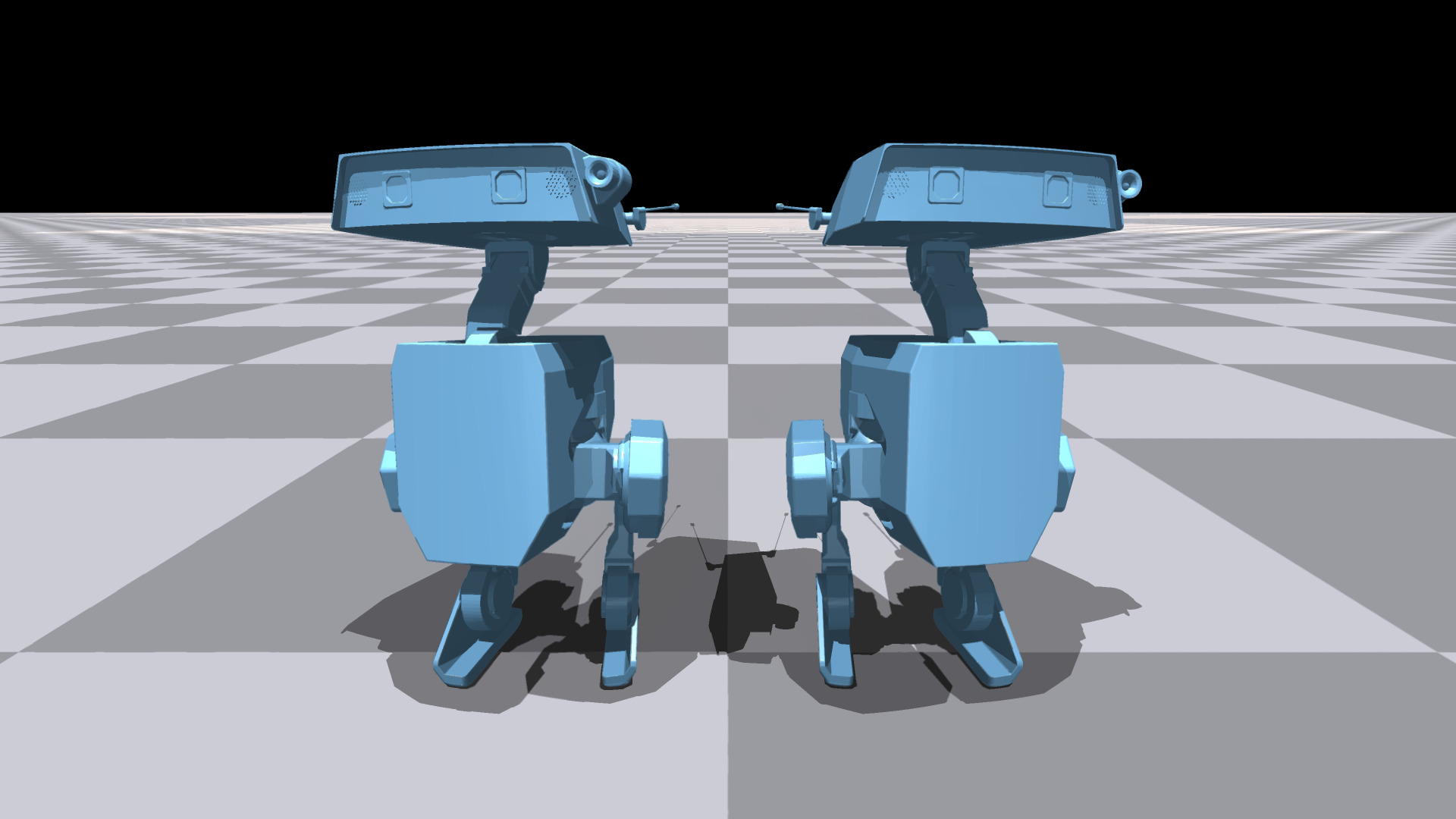} & 
\includegraphics[trim={250 0 250 0},clip,width=0.187\linewidth]{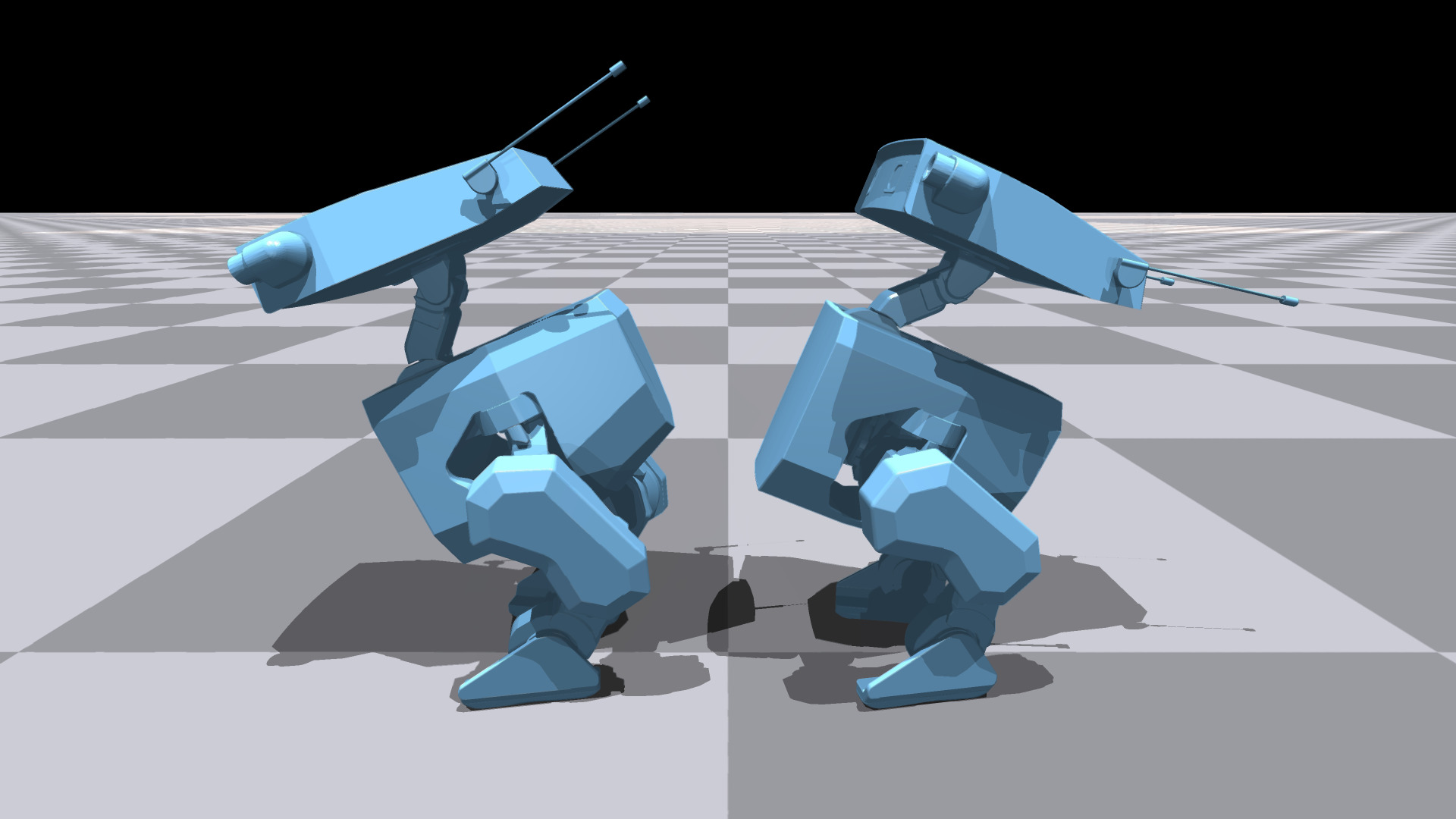} & 
\includegraphics[trim={250 0 250 0},clip,width=0.187\linewidth]{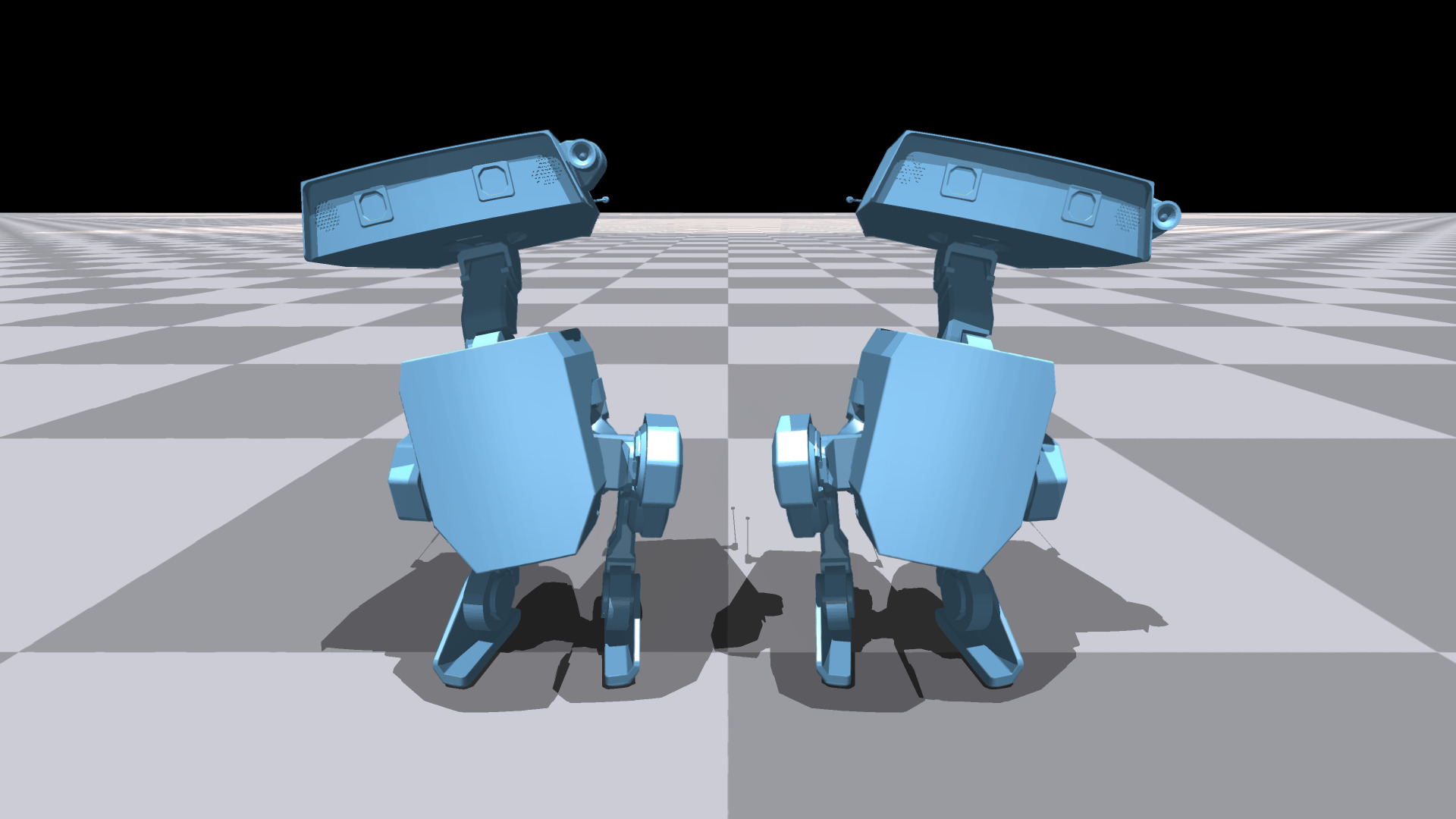} \\ 
\end{tabular}
}
\caption{Path frame illustrations for standing (top-left), and a top-view during walking (bottom-left). During standing, the path frame converges towards the average position and heading of the feet. During walking, the path frame is integrated according to the path velocity commands. The horizontal torso trajectory, shown in blue, will generally sway around the path frame to shift weight between the feet. The other images show the kinematic reference pose for the perpetual standing motion at minimum and maximum value for head commands (top) and torso commands (bottom). The input dimension from left to right are: up-down, yaw, pitch, roll.}
\label{fig:path-frame-control-inputs}
\end{figure*}

\section{Reinforcement Learning}

As outlined in Fig.~\ref{fig:overview}, we control the robot with multiple policies that we condition on a time-varying control input $\vect{g}_t$. At each time step, the agent produces an action $\vect{a}_t$ according to a policy, $\pi(\vect{a}_t | \vect{s}_t, \phi_t, \vect{g}_t)$, provided with the observable state $\vect{s}_t$ and optional conditional inputs $\phi_t$ and $\vect{g}_t$, where $\phi_t$ is a phase signal. During training, the environment then produces the next state, $\vect{s}_{t+1}$, updates the phase signal, and returns a scalar reward $r_t = r(\vect{s}_t, \vect{a}_t, \vect{s}_{t+1}, \phi_t, \vect{g}_t)$. We reward the close imitation of artist-specified kinematic reference motions, and also dynamic balancing.

To bring structure into the broad range of possible character performances, we use differences in their temporal properties to define three motion types:
\begin{itemize} 
    \item \textit{Perpetual} motions do not have a clear start and end. The robot maintains balance and responds to the measured state and a continuous stream of control inputs. 
    \item \textit{Periodic} motions are characterized by a periodic phase signal which is passed to the policy. In this mode, the phase signal cycles indefinitely. 
    \item \textit{Episodic} motions have a predefined duration. Policies receive a monotonically-increasing phase signal. Once the motion ends, a transition to a new motion is forced.
\end{itemize}

For a minimally complete walking character, it is advisable to train at least one policy for perpetual motion and one to imitate a periodic motion. Episodic policies are optional, but well-suited to have the robot express distinct emotions. We treat the training of each policy as a separate RL problem and randomize the control inputs, $\mathbf{g}_t$, over their full range during training. This enables us to control the robot with arbitrary control inputs within their target range, and to switch between policies on the fly during runtime. 

For the presented character, we trained one \textit{perpetual} policy for standing while controlling the head and torso, one \textit{periodic} policy for walking with separate head control, and several \textit{episodic} policies that are each trained to imitate a single animation sequence. In addition, we perturb the simulation model with randomized disturbances and model parameters. In this way, the agent is able to robustly perform the intended motions under a broad distribution of states and user inputs. In the following, we describe these elements in more detail.  

\subsection{Animation Input}
\label{sec:animation_input}

We interface with animation content by extracting kinematic motion references that define the character's time-varying target state,
\begin{equation}
\vect{x}_t = (\vect{p}_t, \vect{\theta}_t, \vect{v}_t, \vect{\omega}_t,  \vect{q}_t, \dot{\vect{q}}_t, c^L_t, c^R_t),
\end{equation}
where $\vect{p}_t$ is the global position of the torso and $\vect{\theta}_t$ its orientation, represented with a quaternion. $\vect{v}_t$ and $\vect{\omega}_t$ are the torso's linear and angular velocity, and $\vect{q}_t$ and $\dot{\vect{q}}_t$ the joint positions and velocities. The indicator variables, $c^L_t$ and $c^R_t$, define the contact state of the character's left and right foot.

For each motion type, we assume that we have access to a generator function $f$ that maps a path frame, $\vect{f}_t$, and an optional phase signal and type-dependent control input to the kinematic target state
\begin{align}
   \vect{x}_t &= f^{\text{perp}} (\vect{f}_t, \vect{g}^{\text{perp}}_t) \\
   (\vect{x}_t, \dot{\phi}_t) &= f^{\text{peri}} (\vect{f}_t, \phi_t,  \vect{g}^{\text{peri}}_t) \\
   \vect{x}_t &= f^{\text{epis}} (\vect{f}_t, \phi_t).
\end{align}
The reference generator for periodic motions additionally outputs the phase rate, $\dot{\phi}_t$, that drives the phase signal. This allows us to vary the stepping frequency during walking as a function of the commands. For episodic motions, the phase rate is determined by the duration of the motion. See Fig.~\ref{fig:path-frame-control-inputs} left for a visualization of the path frame that we use to parameterize our motion references.

For the perpetual standing motion of our specific character, we found that a control input that consists of head and torso commands provides a versatile interface as we illustrate in Fig.~\ref{fig:path-frame-control-inputs} with the limits of each individual head (top row) and torso command (bottom row): We command the head's height and orientation relative to a nominal configuration with a height and orientation offset, $\Delta h^\text{head}_t$ and $\Delta \vect{\theta}^{\text{head}}_t$. The torso position and orientation is commanded with a height, $h^\text{torso}_t$, and orientation, $\vect{\theta}^{\text{torso}}_t$, represented with ZYX-Euler angles, in path frame coordinates
\begin{equation}
    \vect{g}^{\text{perp}}_t = (\Delta h^\text{head}_t, \Delta \vect{\theta}^{\text{head}}_t, h^\text{torso}_t, \vect{\theta}^{\text{torso}}_t).
    \label{eq:g_perpetual}
\end{equation}

For the periodic motion of our droid-like character, we use the same head commands, and use a 2D velocity vector, $\vect{v}^\mathcal{P}_t$, and angular rate, $\omega^\mathcal{P}_t$, both expressed in path frame coordinates
\begin{equation}
    \vect{g}^{\text{peri}}_t = (\Delta h^\text{head}_t, \Delta \vect{\theta}^{\text{head}}_t, \vect{v}^\mathcal{P}_t, \omega^\mathcal{P}_t).
    \label{eq:g_periodic}
\end{equation}

Note that for periodic motion we assume the head commands to be offsets relative to a nominal head motion that the animator specifies as part of the periodic walk cycle. This set of commands puts artists in control of the velocity-dependent lower-body motion while the corresponding references for the neck-head assembly can be adapted to have the robot look in a particular direction during locomotion.

The path frame plays a fundamental role in maintaining consistency during motion transitions. Each artist-designed motion is stored in path coordinates and mapped to world coordinates based on the path frame state according to the generators $f$. During standing, the path frame slowly converges to the center of the two feet. During walking, the next frame is computed by integrating the path velocity commands. This is visualized in Fig.~\ref{fig:path-frame-control-inputs} left. For episodic motions, the path frame trajectory relative to the starting location is part of the artistic input. Finally, to prevent excessive deviation from the path, $\vect{f}_t$ is projected to a maximum distance from the current torso state.

Our processing is agnostic to the tool or technique that an artist uses to generate kinematic reference motion and a detailed description of our specific implementation is beyond the scope of this paper. Hence, we provide a high-level summary here. To generate perpetual references, we use inverse dynamics to find a pose that satisfies the commands $\vect{g}^{\text{perp}}_t$ and solves for optimal values for the remaining degrees of freedom such that the center of pressure is in the middle of the support polygon. For periodic walking motions, an artist provides reference gaits at several walking speeds, defined as task space trajectories for torso and end-effectors. These gait samples are procedurally combined~\cite{hopkins2024interactive} 
into a new gait based on the commands $\vect{g}^{\text{peri}}_t$. To translate the task-space reference to whole-body references, first a model predictive controller similar to~\cite{wieber2006trajectory} is used to plan the desired center of mass and center of pressure. Subsequently, these references are then tracked by an inverse dynamics controller similar to~\cite{koolen2016design} to obtain whole-body trajectories. This processing is part of an interactive tool. The animator can directly preview the resulting gait and fine-tune the task-space references accordingly. The episodic motions are generated in Maya~\cite{maya}.

To prevent the rate at which we can generate reference motions from slowing down training, we densely sample the reference generators and implement the reference look-up during RL as interpolation of these samples. 

\subsection{Reward}

The reward function combines a motion-imitation rewards with additional regularization and survival rewards,
\begin{equation}
    r_t = r^{\text{imitation}}_t + r^{\text{regularization}}_t + r^{\text{survival}}_t
\end{equation}
Following previous work on tracking-based imitation learning~\cite{lee2010datadriven, peng2018deepmimic}, we compute $r^{\text{imitation}}_t$ directly by comparing the simulated to the target pose of the character. The agent receives additional rewards if the foot contact states match the reference states. To mitigate vibrations and unnecessary actions, we apply regularization rewards that penalize joint torques and promote action smoothness. A final survival reward provides a simple objective that motivates the character to stay alive and prevents it from seeking early termination of the episode at the beginning of training. We apply early termination when either head or torso are in contact with the ground and also if we detect a self-collision between the head and torso. A detailed description of the weighted reward terms are provided in Tab.~\ref{tab:rewards} where we use a hat for target state quantities from the reference pose $\mathbf{x}_t$ and omit the current time index $t$ but add spatial indices where necessary.

\begin{table}[tb]
\begin{center}
    \caption{Weighted Reward Terms}
    \label{tab:rewards}
    \begin{tabular}{l | l | l}\toprule 
    \textbf{Name}                  &  \textbf{Reward Term} & \textbf{Weight}\\
    \midrule
    \multicolumn{3}{c}{\textit{Imitation}}  \\
    \midrule
    Torso position xy          &  $\exp\left(-200.0 \cdot \lVert \vect{p}_{x,y} - \hat{\vect{p}}_{x,y} \rVert ^2 \right)$ & $ 1.0 $ \\
    Torso orientation        &  $\exp\left(-20.0 \cdot \lVert \vect{\theta} \boxminus \hat{\vect{\theta}} \rVert ^2\right)$ & $ 1.0 $ \\
    Linear velocity xy  &  $\exp\left(-8.0 \cdot \lVert \vect{v}_{x,y} - \hat{\vect{v}}_{x,y} \rVert ^2\right)$ & $ 1.0 $ \\
    Linear velocity z &  $\exp\left(-8.0 \cdot \left( v_z - \hat{v}_z \right) ^2\right)$ & $ 1.0 $ \\
    Angular velocity xy  &  $\exp\left(-2.0 \cdot \lVert \vect{\omega}_{x,y} - \hat{\vect{\omega}}_{x,y} \rVert ^2 \right)$ & $ 0.5 $ \\
    Angular velocity z   &  $\exp\left(-2.0 \cdot \left( \omega_z - \hat{\omega}_z \right) ^2\right)$ & $ 0.5 $ \\
    Leg joint positions            &  $-\lVert \vect{q}_l - \hat{\vect{q}}_l \rVert ^2 $ & $ 15.0 $ \\
    Neck joint positions           &  $-\lVert \vect{q}_n - \hat{\vect{q}}_n \rVert ^2 $ & $ 100.0 $ \\
    Leg joint velocities            &  $-\lVert \dot{\vect{q}}_l - \hat{\dot{\vect{q}}}_l \rVert ^2 $ & $ 1.0 \cdot 10^{-3} $ \\
    Neck joint velocities           &  $-\lVert \dot{\vect{q}}_n - \hat{\dot{\vect{q}}}_n \rVert ^2 $ & $ 1.0 $ \\
    Contact                 &   $ \sum_{i\in\{L,R\}} \text{I}\left[c_i =\, \hat{c}_i\right]$ & $ 1.0 $ \\
    \midrule
    \multicolumn{3}{c}{\textit{Regularization}}  \\
    \midrule
    Joint torques                 &  $-\lVert \vect{\tau} \rVert ^2 $ & $1.0 \cdot 10^{-3}$ \\
    Joint accelerations           &  $-\lVert \ddot{\vect{q}} \rVert ^2 $ & $2.5 \cdot 10^{-6}$ \\
    Leg action rate         &  $-\lVert \vect{a}_l - \vect{a}_{t-1,l} \rVert ^2 $ & $ 1.5 $ \\
    Neck action rate        &  $-\lVert \vect{a}_n - \vect{a}_{t-1,n} \rVert ^2 $ & $ 5.0 $ \\
    Leg action acc.      &  $-\lVert \vect{a}_l - 2\vect{a}_{t-1,l} + \vect{a}_{t-2,l} \rVert ^2 $ & $ 0.45 $ \\
    Neck action acc.     &  $-\lVert \vect{a}_n - 2\vect{a}_{t-1,n} + \vect{a}_{t-2,n} \rVert ^2 $ & $ 5.0 $ \\
    \midrule
    \multicolumn{3}{c}{\textit{Survival}}  \\
    \midrule
    Survival                &  $1.0$ & $ 20.0 $ \\
    \bottomrule
    \end{tabular}
\end{center}
\end{table}

\subsection{Policy}

Our policy actions, $\vect{a}_t$, are joint position setpoints for proportional-derivative (PD) controllers. In addition to an optional motion-specific phase and control command, our policies receive a state
\begin{equation}
    \vect{s}_t = (\vect{p}^\mathcal{P}_t, {\vect{\theta}}^\mathcal{P}_t, {\vect{v}}^\mathcal{T}_t, {\vect{\omega}}^\mathcal{T}_t,  \vect{q}_t, \dot{\vect{q}}_t, \vect{a}_{t-1}, \vect{a}_{t-2} )
\end{equation}
as input. To make the state and policy invariant to the robot's global location, we represent the torso's horizontal (xy-plane) position, $\vect{p}^\mathcal{P}_t$, and orientation, $\vect{\theta}^\mathcal{P}_t$, in path frame coordinates, and the torso's linear and angular velocities, $\vect{v}^\mathcal{T}_t$ and $\vect{\omega}^\mathcal{T}_t$, in body coordinates. We also append the joint positions, joint velocities, and the actions at the previous two time-steps. All policies are trained with PPO~\cite{schulman2017proximal}. The policy architectures and additional RL details are described in App.~\ref{sec:policy_structure}. 

\subsection{Low-level Control}
The policy outputs actions at a rate of \SI{50}{\hertz}, while our actuator communication operates at \SI{600}{\hertz}. To bridge this gap, we perform a first-order-hold, i.e. a linear interpolation of the previous and current policy action, followed by a low-pass filter with a cut-off frequency of \SI{37.5}{\hertz}. The low-level controller also implements the path frame dynamics and phase signal propagation described in Sec.~\ref{sec:animation_input}. These low-level control aspects are identically implemented in the RL and runtime environments.

\subsection{Simulation}

From the CAD model of the robot, we derive a simulation model that accurately describes the physics of the robot, its actuators, and the interaction of the robot with the environment. The rigid body dynamics of the robot is simulated with Isaac~Gym~\cite{makoviychuk2021isaac}. To accurately describe its full dynamics, we add custom actuator models~\cite{hwangbo2019learning,tan2018sim}. The employed models are derived from first principles with parameters obtained from system identification experiments of the individual actuators (see App.~\ref{sec:actuator_model}). We randomize the parameters of our actuator models within their experimentally observed range. We also add noise to the state that the policy receives, and randomize mass properties and frictional coefficients. In addition to this domain randomization, we apply random disturbance forces and torques on the torso, head, hips, and feet of the robot. During training our walking policy, we additionally randomize the terrain.

\section{Runtime}

After offline training, the weights of the neural control policies are frozen and the policy networks are deployed onto the onboard computer of the robot. Instead of interacting with a simulator, the deployed policies and low-level controllers interface with the robot hardware and a standard estimator of the state $\mathbf{s}_t$, which fuses IMU and actuator measurements~\cite{hartley2020contact}.

The proposed runtime system (compare with Fig.~\ref{fig:overview}) enables an operator to puppeteer the character using an intuitive remote control interface. The \textit{Animation Engine} maps the associated puppeteering commands (including policy switching, triggered animation events, and joystick input) to policy control commands, show function signals, and audio signals. A complete list of puppeteering commands is provided in App.~\ref{sec:interface}. 

We differentiate between artist-specified motions that are used in imitation objectives during RL training, and artist-defined animations that are part of an animation library that a puppeteer interfaces with during runtime. The output of the animation engine is an animation target state for a robot that we then use to form control inputs, $\vect{g}^{\text{perp}}_t$ and $\vect{g}^{\text{peri}}_t$, for our policies. 

\subsection{Perpetual \& Periodic Motions}

\begin{figure}[t]
    \centering
    \includegraphics[trim={250 660 1015 180},clip,width=\linewidth]{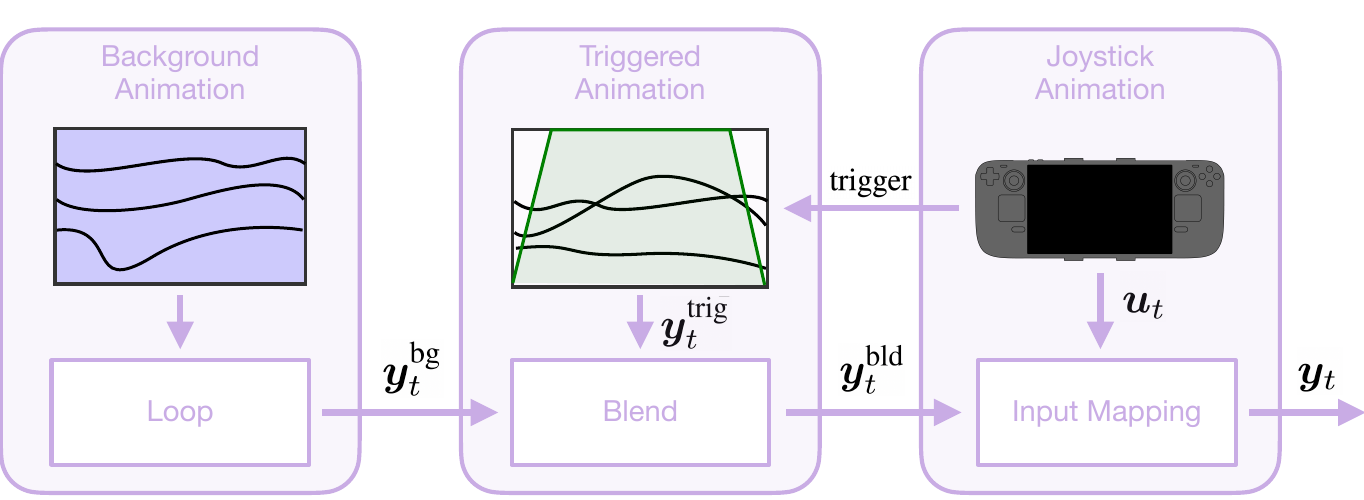}
    \caption{The animation engine procedurally generates the animation command, $\vect{y}_t$, based on three layers: background animation, triggered animations, and animations derived from joystick inputs. A triggered animation is blended in and out as illustrated by the green curve. In contrast, the background animation remains continuously active.}
    \label{fig:animation_engine}
\end{figure}

\begin{table}
\begin{center}
    \caption{Show Function Parameters}
    \label{tab:show-functions}
    \begin{tabular}{l|l|l}
    \toprule
    \textbf{Function Parameters}   & \textbf{Dimensionality} & \textbf{Units} \\  
    \midrule
    Antenna positions & $2 \times 1$ & [\SI{}{\radian}] \\
    Eye colors & $2 \times 3$ & [RGB] \\
    Eye radii & $2 \times 1$ & [\SI{}{\percent}] \\
    Head lamp brightness & 1 & [\SI{}{\percent}] \\
    \bottomrule
    \end{tabular}
\end{center}
\end{table}

During standing and walking, show function and policy commands are computed by combining event-driven animation playback with live puppeteering. To procedurally generate animation states, we define the robot configuration, $\vect{c}_t = ({\vect{p}}^\mathcal{P}_t,{\vect{\theta}}^\mathcal{P}_t, \vect{q}_t)$, from which we extract control inputs. We also define an extended animation command, $\vect{y}_t = (\vect{\nu}_t, \vect{c}_t)$, where $\vect{\nu}_t$ represents all show function commands as summarized in Tab.~\ref{tab:show-functions}. Fig.~\ref{fig:animation_engine} provides a high-level diagram of the proposed animation pipeline which computes a target output, $\vect{y}_t$, by combining three functional animation layers:

\textit{1) Background Animation:} This layer relies on looped playback of a periodic background animation, $\vect{y}^\text{bg}_t$, that is always visible in the absence of additional input. The background animation conveys a basic level of activity that includes intermittent eye-blinking and antenna motion.

\textit{2) Triggered Animations:} This layer blends operator-triggered animation clips on top of the background animation. We select them from a library of artist-specified clips and map them to buttons on the remote control. For our specific character, they range from simple \emph{yes}-\emph{no} animations to complex \emph{scan} clips. Representing the current state of a triggered animation with $\vect{y}^\text{trig}_t$, we blend the background and triggered targets,
\begin{align}
\vect{\nu}^\text{bld}_t &= \; (1 - \beta) \vect{\nu}^\text{bg}_t + \beta \vect{\nu}^\text{trig}_t \\
\vect{c}^\text{bld}_t &= \text{interp}( \vect{c}^\text{bg}_t, \vect{c}^\text{trig}_t, \alpha ) ,
\end{align}
where the configuration interpolation is linear for position and joint angles, and uses \textit{slerp} for body orientation. $\beta$ and $\alpha$ are blend ratios that vary as a function of playback time. Both ratios ramp linearly from 0 to 1 for a given duration at the beginning of an animation and back to 0 over the same duration prior to the end of the animation. We use the durations $T_\beta = \SI{0.1}{\second}$ and $T_\alpha = \SI{0.35}{\second}$, such that the facial expressions associated with the show functions naturally blend faster than the body animation.

\begin{figure}[tb]
\centering
\includegraphics[width=\linewidth]{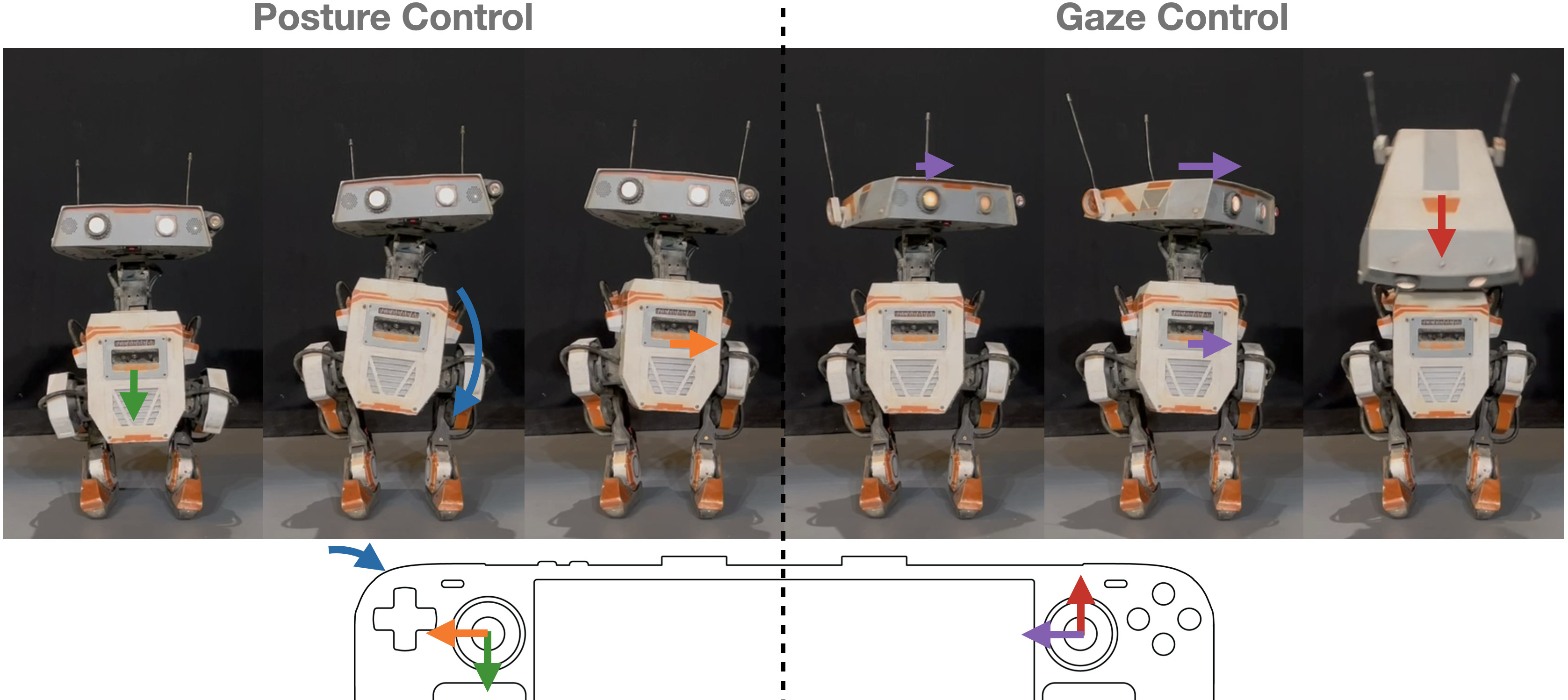}
\caption{A selection of joystick commands during standing. Posture control (left), moves the torso without affecting the gaze. Gaze control (right), changes the gaze primarily by moving the head, but also commands additive torso movement to extend the range. See Appendix~\ref{sec:interface} for details.}
\label{fig:commands}
\end{figure}

\textit{3) Joystick Animation.} The final layer transforms the blended animation state, $\vect{y}^\text{bld}_t$, based on joystick input from the puppeteer. Let $\vect{u}$ represent the joystick axes, triggers, and button modifiers streamed from the controller. While standing, the target robot configuration is computed as
\begin{equation}
\vect{y}_t = \mathcal{J}^\text{perp} \left(\vect{y}^\text{bld}_t, \vect{u}_t \right ),
\end{equation}
where $\mathcal{J}^\text{perp}$ is a non-linear mapping that modifies the current animation state based on the commanded inputs. The joystick axes are mapped to additive offsets to the animated head and torso pose contained in $\vect{c}^\text{bld}_t$, modifying the robot's gaze and posture while idling or executing a triggered animation. Illustrative examples are provided in Fig.~\ref{fig:commands}.

While walking, the target robot configuration is computed using a similar mapping,
\begin{equation}
(\vect{y}_t, \vect{v}^{\mathcal{P}}_t, \omega^{\mathcal{P}}_t) = \mathcal{J}^\text{peri}\left(\vect{y}^\text{bld}_t, \vect{u}_t \right ),
\end{equation}
that additionally produces path velocity commands for the periodic policy. For ease of use, we maintain identical gaze controls; however, the joystick axes used for posture control are remapped to forward, lateral, and turning velocity commands during walking. We also introduce show function modulation based on the path velocity. As the robot reaches top speed, the antennas duck back and the eye radii narrow to convey exertion during fast walking.

Once the animation output, $\vect{y}_t$, has been computed, the show functions are controlled directly with $\vect{\nu}_t$, while the policy command signals are derived from $\vect{c}_t$. For the head, we compare $\vect{c}_t$ to the nominal configuration of the robot and extract the relative head commands $\Delta h^\text{head}_t$ and $\Delta \vect{\theta}^{\text{head}}_t$. During standing, the torso height and orientation in $\vect{g}^{\text{perp}}_t$ are extracted directly from the target configuration. While walking, the lower body motion is determined entirely by the periodic policy and commanded path velocities. Note that in both cases, we ignore the leg joint positions, which are not part of the policy inputs.

\subsection{Episodic Motions}

When an episodic motion is triggered, the animation engine initiates a transition to the corresponding policy and triggers an associated animation clip that syncs the appropriate show function animation, similar to the triggered animation layer during perpetual and periodic motions. There is no additional user input until the episodic motion finishes.

\subsection{Audio Engine}

An onboard audio engine processes and mixes all audio on the robot. A message-based interface enables the operator to trigger short sound clips, e.g. vocalizations, at any time while puppeteering. When an animation or episodic motion has associated audio, the animation engine relays a synchronous playback command to the audio engine. We also support sound effects which are modulated by robot actuator speeds, for creating artificial gear sounds.

\section{Results}
   
We first evaluate the performance and robustness of the individual control policies on which the rest of the control stack builds. Afterward, we show how the animation engine translates user commands into policy control signals and ultimately turns the technical capabilities of the system into a compelling character. Finally, we demonstrate and discuss the deployment of the full system.

\subsection{Evaluating Control Policies}

\textit{Standing:} Each policy input corresponds to the range of motion of one controllable dimension, e.g. the torso yaw. This enables expressive motion during standing, including direct control of the torso. The accompanying video shows the robot going through the full range of each policy input. 

\begin{figure}[tb]
    \centering
    \includegraphics[trim={0 7 0 7},clip,width=0.9\linewidth]{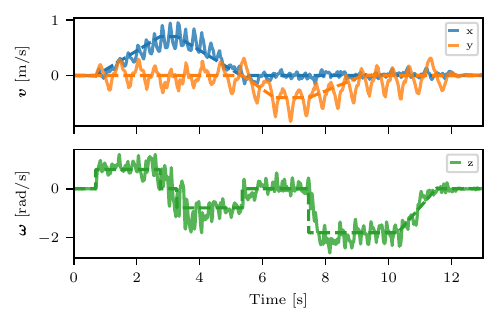}
    \caption{Commanded path velocities (dashed) and measured torso velocities (solid) in path frame.}
    \label{fig:walk_velocities}
\end{figure}

\textit{Walking:} To evaluate walking performance, we demonstrate that the system accurately tracks commanded walking velocities. The walking style has a maximum longitudinal velocity of \SI{0.7}{\meter\per\second}, lateral velocity of \SI{0.4}{\meter\per\second}, and turning rate of \SI{1.8}{\radian\per\second}. Fig.~\ref{fig:walk_velocities} shows the estimated torso velocity of the robot in path frame compared to the commanded walking velocities. The robot is responsive and follows all commands closely.

\begin{figure}[tb]
    \centering
    \includegraphics[trim={0 7 0 7},clip,width=0.9\linewidth]{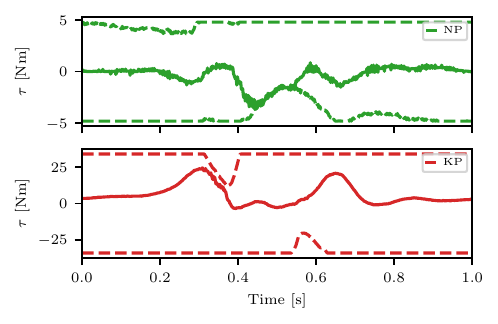}
    \caption{Measured joint torques (solid) and velocity-dependent torque limits (dashed) computed by the actuator model for the measured joint velocities, during the episodic ``jump'' motion. The top plot shows the neck pitch (NP) actuator and the bottom plot shows the left knee pitch (KP) actuator.}
    \label{fig:jump_torques}
\end{figure}

\textit{Episodic Policies:} In the video we show several examples of episodic motions, namely a ``happy dance'', ``excited motion'', ``jump'', and ``tantrum''. They demonstrate a diverse set of motions and have a high level of coordination between joints, at a level best achieved with a specialized policy. Fig.~\ref{fig:jump_torques} shows the torque of both the neck pitch and left knee pitch joints during the jump motion, together with their velocity-depended torque limits as predicted by the actuator model. It can be seen that while pushing off, the robot reaches the torque limits of the knees, which rapidly decrease due to increasing joint velocities. During the jump, the robot makes an upward pitching movement with the head, causing the actuator in the head to also reach its limit. Tracking errors for these and motions from previous sections are reported in Tab.~\ref{tab:tracking_errors}.

\begin{table}[tb]
\centering
\caption{Mean Absolute Tracking Error (MAE) of Joint Positions}
\begin{tabular}{ll|l}
\toprule
Type & Name   & MAE [\SI{}{\radian}]  \\ \midrule
\textit{Perpetual} & Standing & 0.035 \\
\textit{Periodic} & Walking  &  0.123     \\
\textit{Episodic} & Excited Motion  &   0.029    \\ 
& Happy Dance     &   0.027    \\ 
& Jump            &   0.043    \\ 
& Tantrum         &   0.032    \\ 
\bottomrule
\end{tabular}
\label{tab:tracking_errors}
\end{table}

\textit{Robustness:} We demonstrate the robot's ability to stabilize under challenging conditions by subjecting it to external pushes and by walking over small obstacles (see video). The policy deviates from the reference trajectory and contact schedule to recover and maintain balance. This highlights the strength of using an RL approach. In contrast, optimization-based approaches struggle to simultaneously plan motion and contact schedules in real-time, and are often constrained to follow the contact reference \cite{wensing2023optimization}.

\begin{figure}[tb]
    \centering
    \includegraphics[trim={0 7 0 7},clip,width=0.9\linewidth]{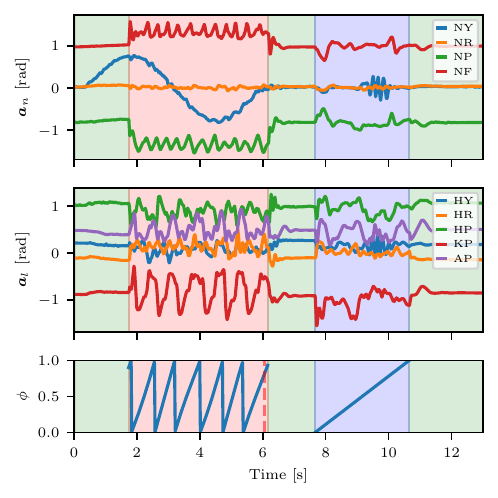}
    \caption{Policy actions across policy transitions during a short motion sequence. The active policy is indicated with the background color of the plot (green: standing, red: walking, blue: episodic). The top plot shows the actions for the neck joints: neck yaw (NY), neck roll (NR), neck pitch (NP), and neck forward (NF). The middle plot shows actions form the left leg: hip yaw (HY), hip roll (HR), hip pitch (HP), knee pitch (KP), and ankle pitch (AP). The bottom plot shows the phase signal when applicable. The dashed red line shows the moment a request to transition from walking to standing is received. The transition is made when the gait reaches the next double support phase.}
    \label{fig:policy_info}
\end{figure}

\textit{Policy Transitions:} Fig.~\ref{fig:policy_info} provides insight into the policy transitions for a short sequence shown in the video. The plots are colored according the active policy. The top two plots show the policy actions, i.e.\, joint position setpoints, for all neck joints and left leg joints. Because each policy receives the two previous actions and smoothness is encouraged during training, the actions remain continuous and the switch of policies is virtually undetectable for an outside observer. 

While it is possible to transition in and out of the walking gait at any point, the transition looks more natural when done at the appropriate, non-zero phase within the gait cycle. When starting to walk, we initialize the phase based on the commanded turning direction: when turning left, we start the phase at a left step, and vice-versa for a right turn. When transitioning out of walk (see dashed line in Fig.~\ref{fig:policy_info}), the policy switch is delayed until the start of the next double support phase. This way the walking policy finishes the swing phase as intended and the standing policy starts with both feet already on the ground.

\subsection{Alternative RL Formulations}
We compare our RL formulation against related alternatives in the literature. First, we compare against an approach that directly tracks the commanded walking velocities with the torso~\cite{hwangbo2019learning}. Specifically, we use the commands as reference in the torso related rewards and set the leg joint position, leg joint velocity, and contact weights to zero. The phase signal is removed from the policy inputs. As shown in the video, this results in a policy that rapidly shuffles the feet. \textit{Foot clearance} and \textit{slip penalties} are commonly used to suppress this behavior~\cite{miki2022learning}. However, they are tedious to tune and the resulting gait remains far from natural. Second, we compare against approaches that provide the policy with a phase signal and corresponding contact reference~\cite{siekmann2021sim,siekmann2021blind}. We add back the phase signal to the inputs and activate the contact reward. As shown in the video, the resulting policy follows the stepping pattern well. However, the effect of directly tracking the velocity commands remains visible, resulting in a stiff and upright torso motion. Finally, we evaluated using the current and future kinematic reference poses as the policy inputs in place of the phase signal~\cite{peng2020learning,li2021reinforcement}, with the rest of the formulation remaining identical to ours. For both the walking and episodic motions this formulation converges to the same reward and produces a visually identical motion. This is not surprising: the phase signal and kinematic reference contain the same information. The benefit of using a phase signal is that we do not need to store and reproduce the reference motions on the robot.

\subsection{Animation Engine}

\begin{figure*}[tb]
{
\footnotesize
\setlength{\tabcolsep}{1pt}
\begin{tabular}{ccc}
\includegraphics[trim={600 420 150 60},clip,width=0.33\linewidth]{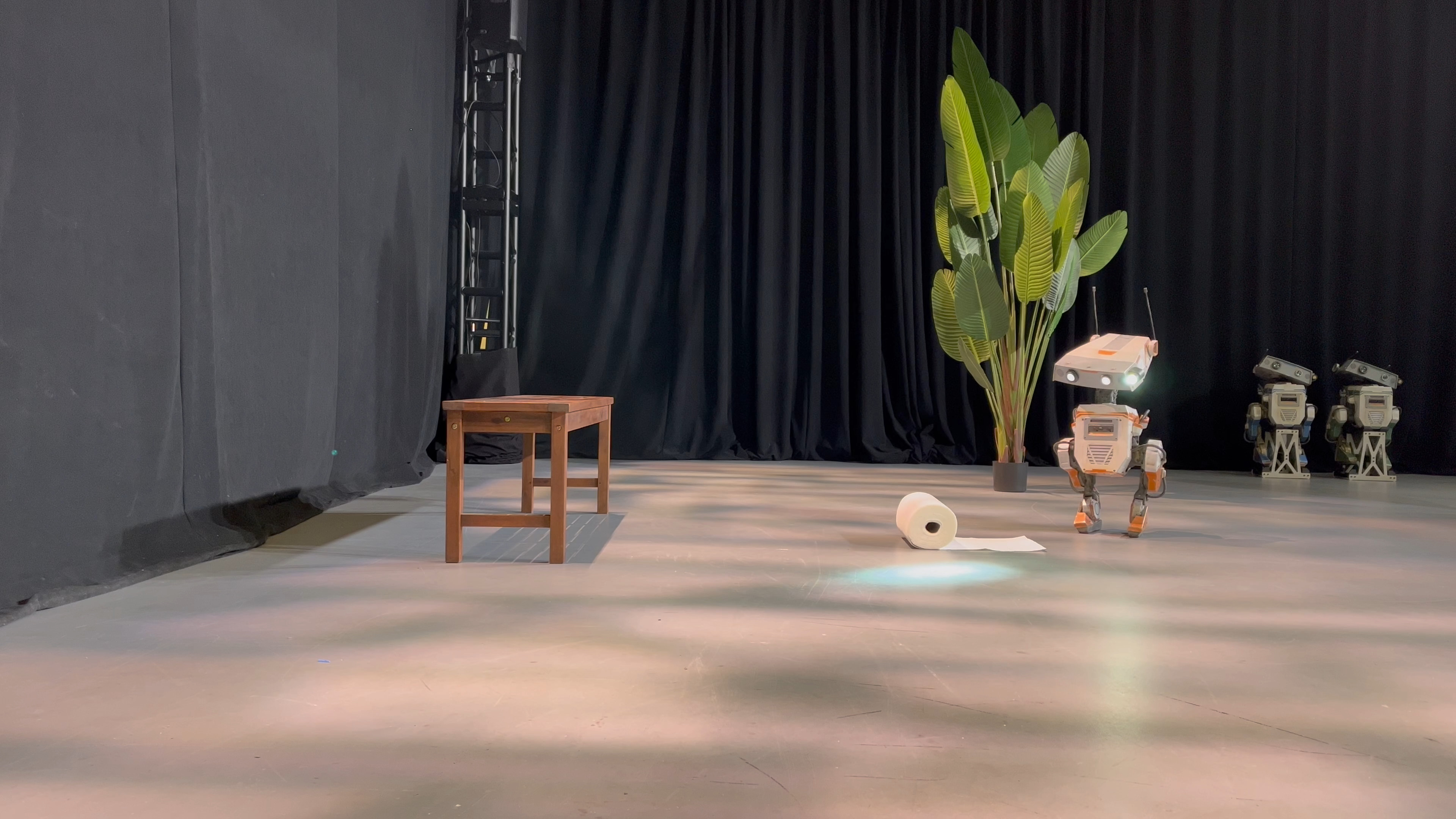} & 
\includegraphics[trim={600 420 150 60},clip,width=0.33\linewidth]{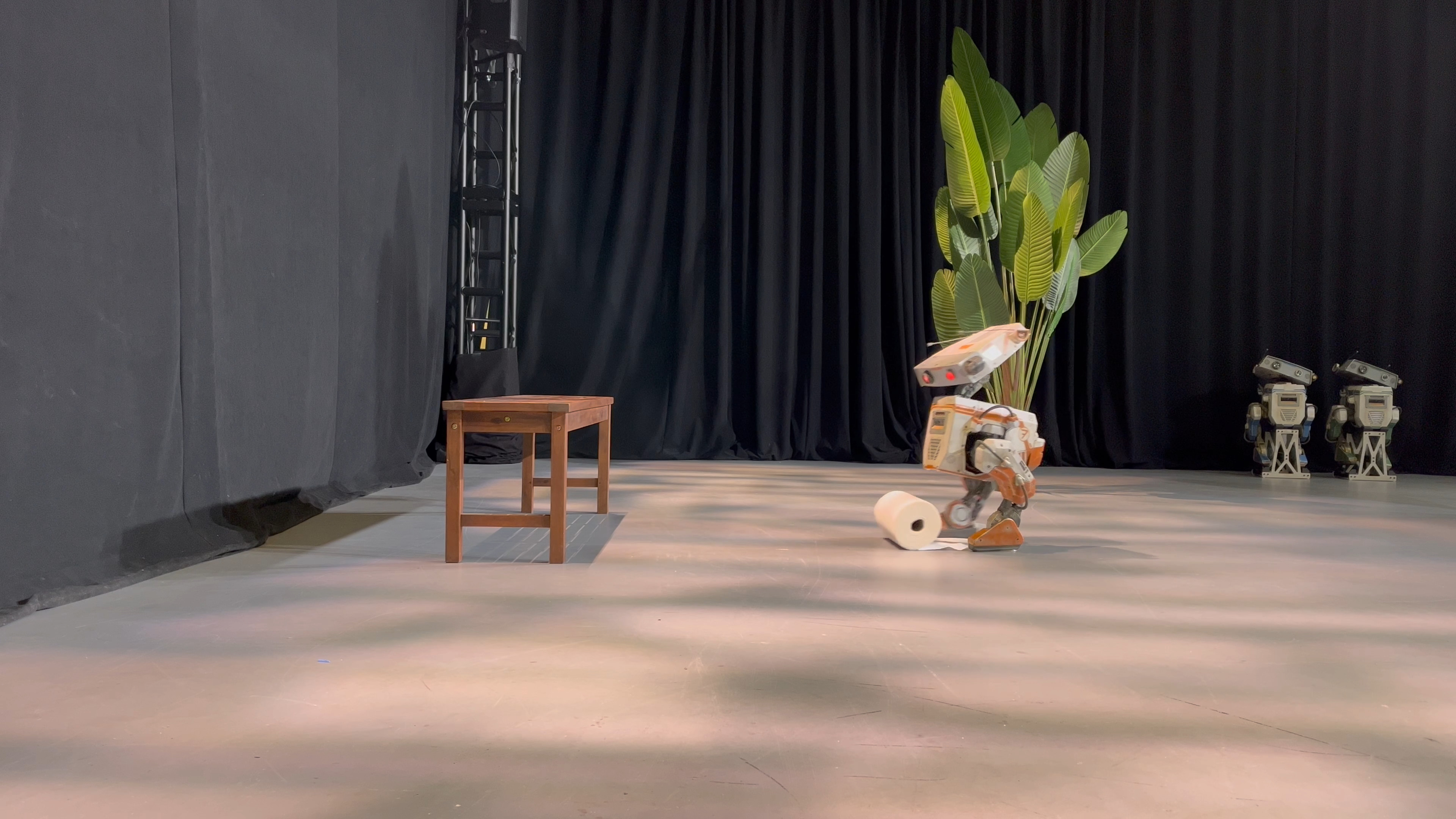} & 
\includegraphics[trim={600 420 150 60},clip,width=0.33\linewidth]{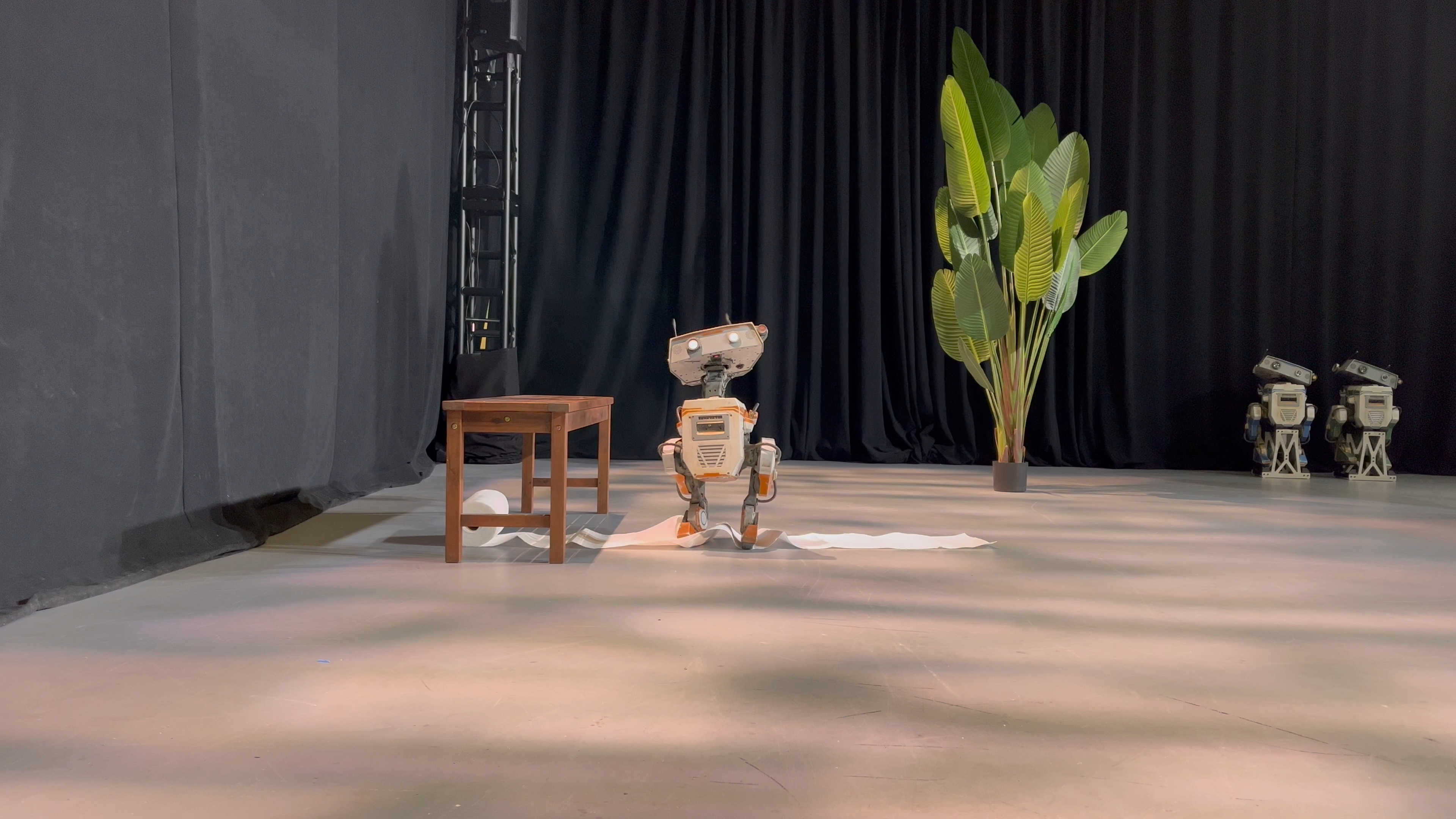}  \\ 
\includegraphics[trim={270 150 0 80},clip,width=0.33\linewidth]{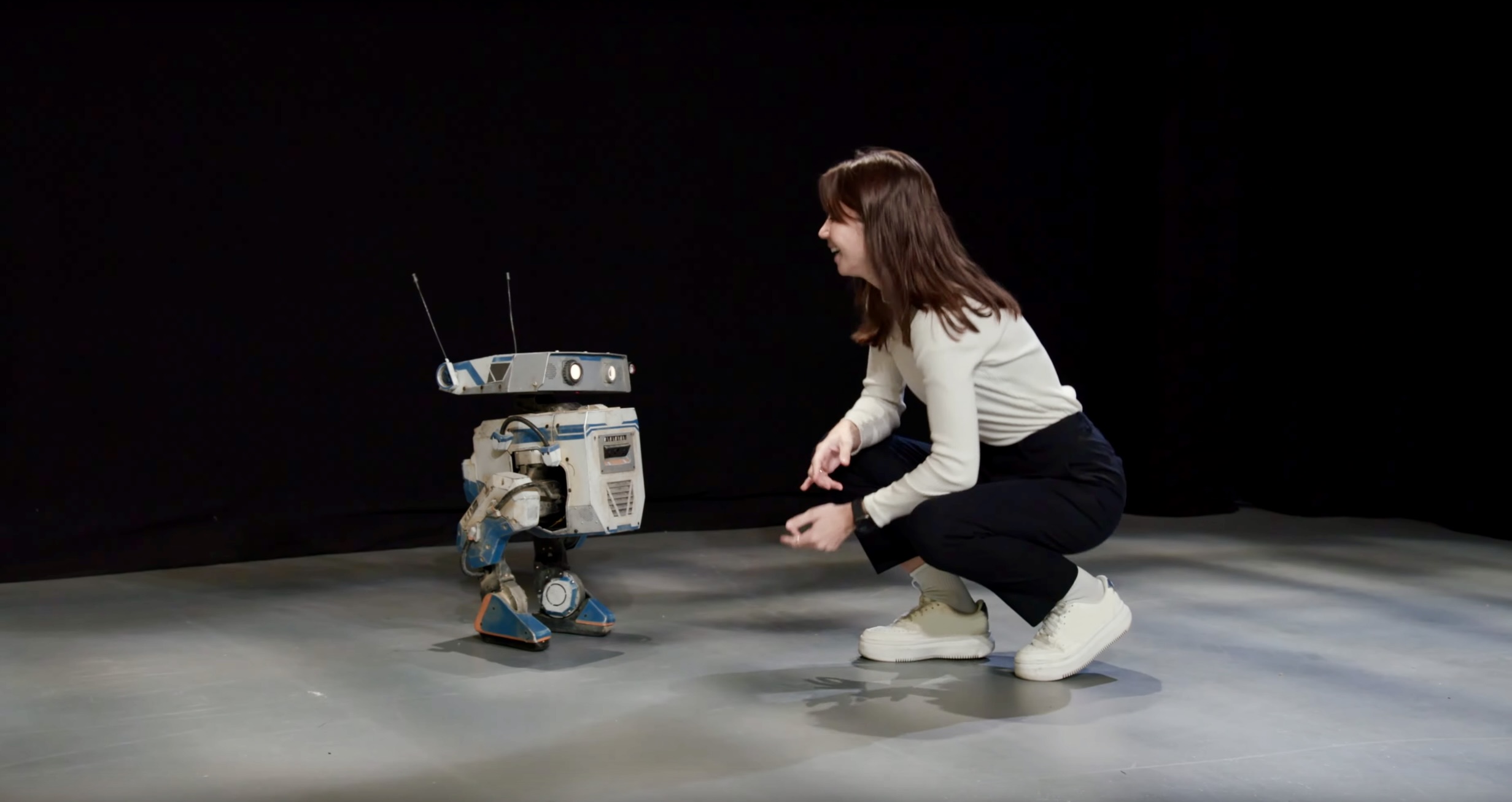} & 
\includegraphics[trim={270 150 0 80},clip,width=0.33\linewidth]{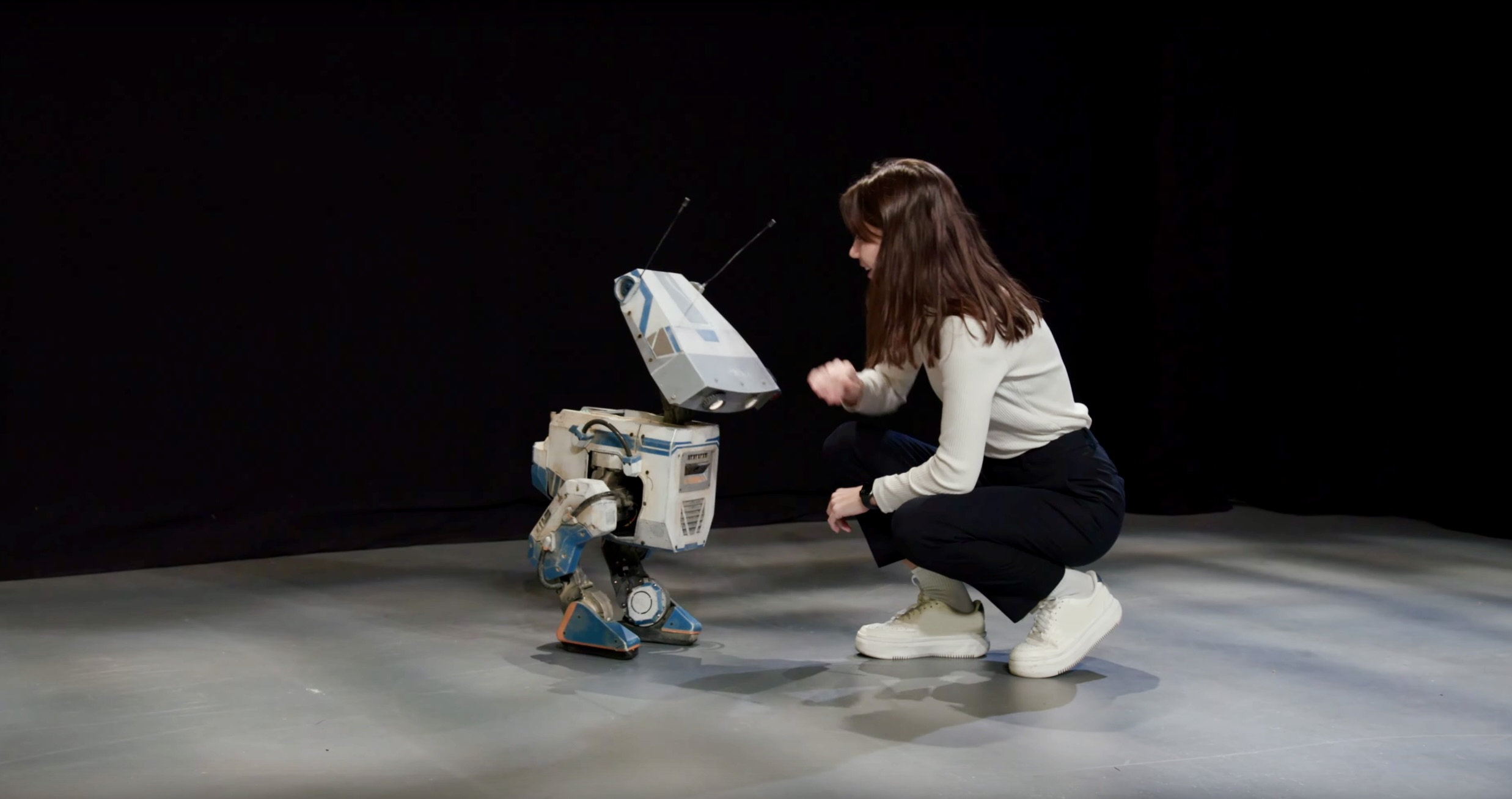} & 
\includegraphics[trim={270 150 0 80},clip,width=0.33\linewidth]{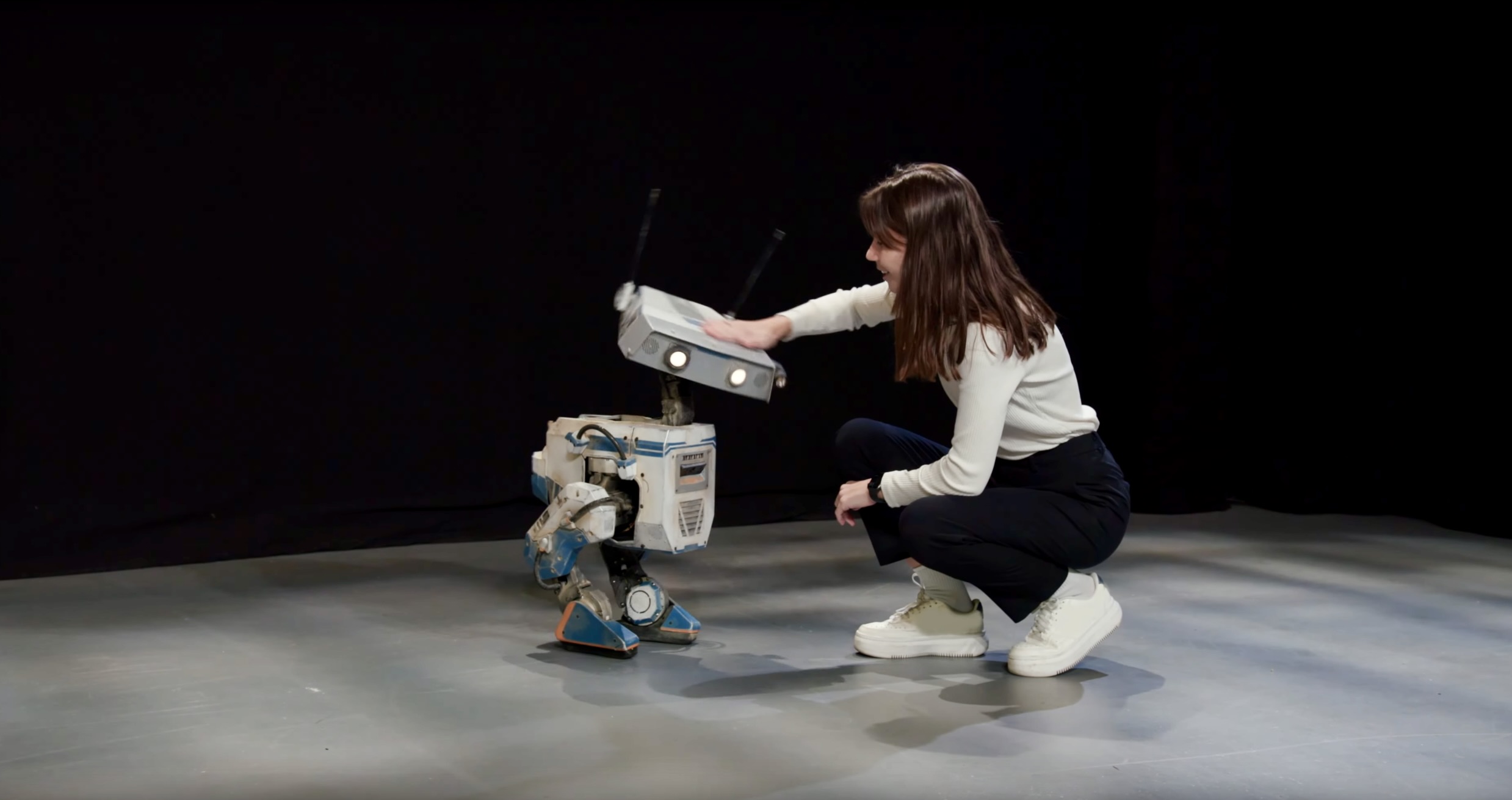}  \\ 
\end{tabular}
}
\caption{(Top) The operator uses the proposed puppeteering interface to act out a scene where the robot discovers a paper roll and decides to kick it under the bench. We make use of the head lamp, triggered head animations, and gaze controls to convey initial curiosity. Episodic policies are triggered following the run-up and kick to convey excitement and celebration. (Bottom) The operator puppeteers the robot to interact with a human. The robot is at first shy, then approaches, extends its head, and is rewarded with a head scratch.}
\label{fig:puppeteering}
\end{figure*}

In the accompanying video, we show the build up of the individual layers of the animation engine. First showing only the background animation, then adding the triggered animation layer, and finally adding the joystick commands. The combination of triggered animation content and joystick-driven targeting enables simple, intuitive, and expressive puppeteering of the robotic character. With direct teleoperation, the operator can deliver a wide range of performances adapted to the robot’s environment. Meanwhile, the operator can trigger animation clips that enable expressions and interactions with subtle or complex movement that cannot be achieved through joystick control alone. As an example, the puppeteer can direct the robot's gaze toward a person posing a question, then trigger a stylized \emph{yes} or \emph{no} animation while simultaneously maintaining eye contact and modifying the body posture. Fig.~\ref{fig:puppeteering} includes stills from the video demonstrating the effectiveness of the proposed approach for puppeteering the robot to act out a short scene and interact with a human.

\textit{Gaze and Posture Control:}
For the remote control interface, we selected a two-joystick device with a layout familiar to operators with video game experience. Designing an appropriate joystick input mapping is challenging due to the high dimensionality of the robot configuration and limited joystick axes. To reduce cognitive load on the puppeteer, the joystick controls were designed to provide independent control of the robot's gaze and body posture during standing. As illustrated in Fig.~\ref{fig:commands}, the left joystick modifies the body posture (torso yaw and pitch), while counter-rotating the head to maintain a fixed gaze. Meanwhile, the right joystick modifies the robot's gaze (head yaw and pitch), while introducing additive rotation of the torso when the neck reaches it's kinematic limits. When the operator directs the robot's gaze to an extreme angle, either yawing to the side or pitching towards it's feet, the body naturally follows the head. The video results demonstrate each control independently, followed by a combined puppeteering sequence. The functional separation allows the puppeteer to more easily direct the robot's line of sight while simultaneously modifying the body posture to convey emotion. The detailed interface described in Appendix~\ref{sec:interface} was continuously refined throughout the project, including the training of several operators without a background in robotics.

\subsection{System Deployment}

At the time of writing, we have conducted several public deployments of the robot with up to three robots simultaneously, resulting in the order of \SI{10}{\hour} of robot runtime without a single fall.
Audiences quickly become captivated by the robotic character and often will not notice the presence of the puppeteer. However, some feedback has also called out the presence of a puppeteer as a distraction or reducing believability of the character. Frequently bystanders assume the robot can perceive its environment: \textit{``can the robot really see me?''}, or \textit{``how does it know what I'm saying?''}.

\section{Conclusion} 
\label{sec:conclusion}

In this work we propose a robot design and control workflow that targets the intricate challenges associated with legged robots for entertainment applications. 
We present a new bipedal robotic character and demonstrate the integration of expressive, artist-directed motions with robust dynamic mobility. Multiple RL policies, trained to imitate artistic motions, and conditioned on low-dimensional input signals, provide a robust foundation on which we build in the animation engine. Together, they allow for real-time show performances through an intuitive operator interface.

Our work has also demonstrated that it is possible to build dynamic legged robots where the kinematics and the mechanical design are driven by a creative target rather than by functional requirements. Taken together with the general formulation of the presented pipeline, our work enables the creation of expressive robot characters outside the typical anthropomorphic or zoomorphic morphologies, paving the way towards more general and fantastical robotic characters.

While the separation into multiple policies has provided us with precise control over the behavior of the robot, it results in training overhead especially when scaling up the amount of episodic motions. Here, we are eager to explore if a single policy can learn several skills with the same level of accuracy as the specialized policies. Finally, there is a natural limit to how many buttons a puppeteer can use effectively. To further expand the expressive capabilities of the character, we see an opportunity for embedding autonomy in the animation engine.

\bibliographystyle{plainnat}
\bibliography{references}

\appendices

\section{Policy Training}
\label{sec:policy_structure}

Each policy is trained for $100\,000$ iterations, which corresponds to a computation time of around $2$ days on a Nvidia RTX 4090. Despite the long training time for the final policy, we note that most of the behavior is learned in the first few iterations. In our experience, similar to related work~\cite{rudin2022learning}, $1500$ iterations or \SI{30}{\minute} are enough to preview the nominal behavior. We use the PPO 
hyperparameters shown in Tab.~\ref{tab:hyperparams_rl}, with and adaptive learning rate~\cite{rudin2022learning}. For all policies, we use three fully connected layers of 512 hidden units and ELU activation functions. For the critic, we use a separate network with the same amount of hidden units. In addition to the policy inputs, the critic receives the simulation state without noise and the randomized friction parameter as privileged information. 

Moreover, we apply fixed transformations, i.e. without learnable parameters, on both the input and output side of the policy. We first normalize all inputs by their expected range. For policies that take phase as an input, we replace the phase with a feature vector derived from the phase. For walking, we add the first two harmonics $(\sin(k \cdot 2\pi \phi), \cos(k \cdot 2\pi \phi))$, $k\in\{1, 2\}$. The first harmonic corresponds to the walking cycle and the second harmonic makes it easier for the policy to learn the head-bob which occurs at twice the gait frequency. For the episodic motions, we use $N=50$ Gaussian basis functions, $\exp\left(-(\phi - \phi_i)^2/ (2 N^2)\right)$, where the $\phi_i$s are equally spaced between $0$ and $1$. These features are highly local in time and allow the policy to form a rich phase-dependent feed-forward signal. At the output, we apply a linear transform to the actions such that $0$ maps to the nominal joint positions of the robot and $1$ to the expected range per joint. Finally, joint position setpoints are clipped to a maximum deviation around the measured joint position. This deviation is chosen large enough such that the maximum actuator torque can still be produced.

We use the same rewards listed in Tab.~\ref{tab:rewards} for all policies, with small modifications for the episodic motions ``excited motion'' and ``jump''. For both motions, we emphasize a key aspect of the motion by increasing selected weights between $\phi_\text{start}$ and $\phi_\text{end}$,
\begin{equation}
    \tilde{w}(\phi) = w_0 + \text{I}\left[\phi_\text{start} < \phi < \phi_\text{end} \right] \cdot w_\text{extra},
\end{equation}
where $\text{I}\left[\cdot\right]$ is an indicator function. For the ``excited motion'', we increase the weight on angular velocity tracking during the part where the robot rapidly shakes its torso. For the ``jump'', we add a reward for torso height and add extra weight on the torso orientation during the jump. Additionally, we increase the contact reward during the jump. Without this weight adjustment, the policy tends to cheat by keeping the toes on the ground.

Details for the applied disturbance forces are given in Tab.~\ref{tab:dist_forces}. All parameters refer to the minimum and maximum magnitude drawn from a uniform distribution. For each specified body, a random force and torque drawn from a uniform distribution per dimension is applied for a random ``on'' duration. Afterward, a random ``off'' duration is selected until the next disturbance is applied. We organize the parameters in three separate categories, but the process is applied independently per body. The forces are gradually introduced according to a linear curriculum over the first $1500$ iterations.

\begin{table}[tb]
\begin{center}
\caption{RL Hyperparameters}
\footnotesize
\begin{tabular}{l|l}
\toprule
\textbf{Param.}      & \textbf{Value}     \\ \midrule
Num. iterations & $100\,000$ \\
Batch size
$(\text{envs.}\times\text{steps})$ & $8192\times24$  \\
Num. mini-batches & $4$  \\  
Num. epochs  &  $5$  \\
Clip range & $0.2$  \\
Entropy coefficient & $0.0$ \\
Discount factor  & $0.99$  \\
GAE discount factor & $0.95$  \\
Desired KL-divergence & $0.01$  \\
Max gradient norm & $1.0$  \\
\bottomrule
\end{tabular}
\label{tab:hyperparams_rl}
\end{center}
\end{table}

\begin{table}[htbp]
\centering
\caption{Disturbance Parameters}
\footnotesize
\begin{tabular}{ll|lll}
\toprule
Param.                          &     &  Short / small & Long / small & Short / large \\ \midrule
Body                          &     & Hips, Feet & Pelvis, Head   &  Pelvis \\
Force [\SI{}{\newton}]   & XY  &  [0.0, 5.0]  & [0.0, 5.0]  &  [90.0, 150.0] \\
                              & Z   &  [0.0, 5.0]  & [0.0, 5.0]  &  [0.0, 10.0] \\
Torque [\SI{}{\newton\meter}]  & XY  &  [0.0, 0.25] & [0.0, 0.25] &  [0.0, 15.0] \\
                              & Z   &  [0.0, 0.25] & [0.0, 0.25] &  [0.0, 15.0] \\
Duration [\SI{}{\second}] & On  &  [0.25, 2.0] & [2.0, 10.0] &  [0.1, 0.1] \\
                              & Off &  [1.0, 3.0]  & [1.0, 3.0] & [12.0, 15.0] \\ \bottomrule
\end{tabular}
\label{tab:dist_forces}
\end{table}

\section{Actuator Model}
\label{sec:actuator_model}

We describe the actuator models that we use to augment the physics simulator. System identification was performed by mounting single actuators on a test-bench that measures output torque. All identified parameters are provided in Tab.~\ref{tab:actuator_model}. 

The proportional derivative motor torque equation for the quasi-direct drives used in this work is computed through,
\begin{equation}
    \tau_{m} = k_{\text{P}} (a - \tilde{q}) - k_{\text{D}} \dot{q},
\end{equation}
where $k_{\text{P}}$ and $k_{\text{D}}$ are gains, $a$ is the joint setpoint, $\dot{q}$ is the joint velocity, and $\tilde{q}$ is the joint position with an added encoder offset, $\epsilon_{q}$,
\begin{equation}
    \tilde{q} = q + \epsilon_{q}.
\end{equation}
This encoder offset is drawn from a uniform distribution, $\mathcal{U}(-\epsilon_{q,\text{max}}, \epsilon_{q,\text{max}})$, at the beginning of each RL episode and accounts for inaccuracy in our joint calibration.

The actuators used in the head do not implement the same PD equation and instead use,
\begin{equation}
    \tau_{m} = k_{\text{P}} (a - \tilde{q}) + k_{\text{D}} \frac{\text{d}}{\text{d}t} (a - \tilde{q}),
\end{equation}
where the derivative term operates on the numerical differentiation of the setpoint error.

Friction in the joint is modelled as,
\begin{equation}
    \tau_{f} = \mu_s \tanh{\left(\dot{q} / \dot{q}_s \right)} + \mu_d \dot{q},
\end{equation}
where $\mu_s$ is the static coefficient of friction, with activation parameter $\dot{q}_s$, and $\mu_d$ is the dynamic coefficient of friction. 

The torque produced at the joint can then be computed by applying torque limits to $\tau_{m}$ and subtracting friction forces,
\begin{equation}
    \tau = \text{clamp}_{\left[\underline{\tau}(\dot{q}), \overline{\tau}(\dot{q})\right]} \left( \tau_{m} \right) - \tau_{f},
\end{equation}
where $\underline{\tau}(\dot{q})$, and $\overline{\tau}(\dot{q})$, are the velocity-dependent minimum and maximum torques of the motor. These limits consists of a constant limit torque for braking and low velocities, $\tau_\text{max}$, and a linear limit ramping down the available torque above an identified velocity, $\dot{q}_{\tau_\text{max}}$. The linear limit crosses the velocity, $\dot{q}_\text{max}$, when the limit torque is \SI{0}{\newton\meter}.

The measured joint position reported by the actuator model contains the encoder offset, a term that models backlash and noise,
\begin{equation}
    \hat{q} = \tilde{q} + 0.5 \cdot b \cdot \tanh{\left(\tau_{m} / \tau_b \right)} + \mathcal{N}\left(0, \sigma_q^2\right),
\end{equation}
where $b$ is the total motor backlash with activation parameter $\tau_b$. $b$ is drawn from a uniform distribution, $\mathcal{U}(b_\text{min}, b_\text{max})$,  at the start of each episode. This term models the effect that a motor moves to the side of the backlash in the same direction as the applied motor torque. For the noise model, we fit a normal distribution with the standard deviation proportional to the motor rotating velocity
\begin{equation}
    \sigma_q = \sigma_{q,0} + \sigma_{q,1} | \dot{q} |.
\end{equation}

Finally, the reflected inertia of the actuator, $I_{m}$, is added to the simulation model. In Isaac Gym, this is done by setting the \textit{armature} value of the corresponding joint. This value is randomized up to a \SI{20}{\percent} offset at the start of each episode.

\begin{table}[htb]
\begin{center}
    \caption{Actuator Gains and Model Parameters}
    \label{tab:actuator_model}
    \footnotesize
    \begin{tabular}{l|l l l l}\toprule 
     & {Unitree}  & {Unitree} & {Dynamixel} \\
    {Param.} & {A1} & {Go1} & {XH540-V150}  & {Units} \\
    \midrule
    $k_\text{P}$  & $15.0$ & $10.0$ & $5.0$ & [\SI{}{\newton\meter\per\radian}] \\
    $k_\text{D}$  & $0.6$ & $0.3$ & $0.2$ & [\SI{}{\newton\meter\second\per\radian}] \\
    \midrule
    $\tau_\text{max}$  & $34.0$ & $23.7$ & $4.8$ & [\SI{}{\newton\meter}] \\
    $\dot{q}_{\tau_\text{max}}$ & $7.4$ & $10.6$ & $0.2$ &  [\SI{}{\radian\per\second}]\\
    $\dot{q}_\text{max}$ & $20.0$ & $28.8$ & $7.0$ &  [\SI{}{\radian\per\second}]\\
    $\mu_s$  & $0.45$ & $0.15$ & $0.05$ & [\SI{}{\newton\meter}] \\
    $\mu_d$ & $0.023$ & $0.016$ & $0.009$ & [\SI{}{\newton\meter\second\per\radian}] \\
    $b_\text{min}$  & $0.005$ & $0.002$ & $0.002$ & [\SI{}{\radian}]  \\
    $b_\text{max}$  & $0.015$ & $0.005$ & $0.005$ & [\SI{}{\radian}]  \\
    $\epsilon_{q,\text{max}}$ & $0.02$ & $0.02$ & $0.02$ &  [\SI{}{\radian}] \\
    $\sigma_{q,0}$  & $ 1.80 \cdot 10^{-4} $ & $  1.89 \cdot 10^{-4} $ & $ 4.31 \cdot 10^{-4} $ & [\SI{}{\radian}]\\
    $\sigma_{q,1}$  & $ 3.61 \cdot 10^{-5}  $ & $ 5.47 \cdot 10^{-5}  $ & $ 2.43 \cdot 10^{-5} $ & [\SI{}{\second}] \\
    $I_{m}$ & $  0.011 $ & $  0.0043 $ & $ 0.0058 $ & [\SI{}{\kilogram\meter\squared}] \\
    \bottomrule
    \end{tabular}
\end{center}
\end{table}

\section{Puppeteering Interface}
\label{sec:interface}

\begin{figure}[bh]
  \centering{}
    \includegraphics{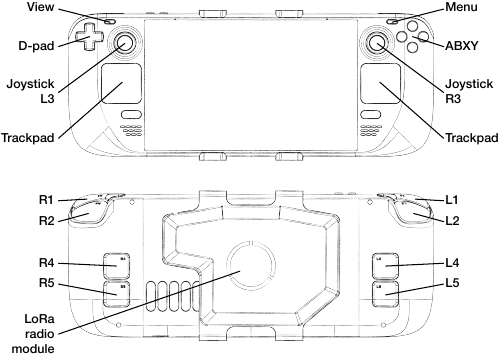}
  \caption{Steam Deck~\cite{steamdeck} layout.}
  \label{fig:steamdeck}
\end{figure}

The button layout of the remote controller used in this work is shown and annotated in Fig.~\ref{fig:steamdeck}. The effect of each button is described in Tab.~\ref{tab:interface}. The simple touchscreen buttons turn on and off the robot. During operation, the screen displays vital system information: battery voltage, motor temperatures, and connection stability.

\begin{figure}[bh]
  \centering
  \includegraphics[width=0.9\linewidth]{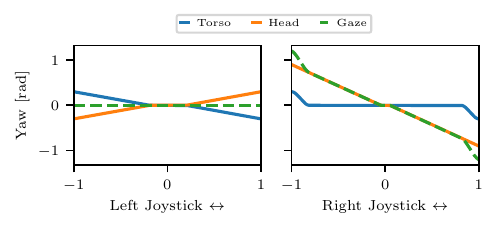}
  \caption{Joystick mapping during standing for the torso and head yaw offset depending on the left and right joystick inputs. The resulting gaze offset is the sum of the torso and local head offset.}
  \label{fig:gamepad_yaw}
\end{figure}

Fig.~\ref{fig:gamepad_yaw} illustrates how the torso and head yaw offsets are computed as a function of the left and right joystick inputs to achieve independent torso movement with the left joystick, and additive gaze and torso coordination with the right joystick. The same principle applies to the pitch and roll directions as summarized in Tab.~\ref{tab:interface}.

\begin{table}[tbh]
\begin{center}
    \caption{Puppeteering Button Mapping}
    \label{tab:interface}
    \begin{tabular}{l c p{0.53\linewidth}}\toprule 
    \textbf{Button} & &  \textbf{Effect} \\\midrule
    Menu & & Trigger a safety mode called \textit{motion stop}. This forces a transition to standing and freezes the joint setpoints with high position gains after waiting \SI{0.5}{\second}. \\
    View & & Slowly move all joints to the default pose. Only available at startup or while in \textit{motion stop}. \\
    D-pad & $\updownarrow$ & Move the head up-down.  \\
          & $\leftrightarrow$ & Roll the head left-right.  \\
    Left Joystick & $\updownarrow$ & During walking: Longitudinal walking velocity. During standing: \textit{Up} pitches the torso forward while the head remains stationary, and \textit{Down} lowers the torso height.  \\
                    & $\leftrightarrow$ & During walking: Turning rate. During standing: Torso yaw while the head remains stationary. \\
                    & L3 & Pressing the left joystick triggers a scanning animation. \\
    Right Joystick & $\updownarrow$ & Pitches the head. During standing, this additionally commands torso pitch. \\
                    & $\leftrightarrow$ & Yaws the head left-right. During standing, the end of the range additionally commands torso yaw. \\
                    & R3 & Pressing the right joystick toggles the audio level. \\
    ABXY & A & Transition to standing. \\
     & B & Fully tuck the neck in, turn off the eyes, and retract the antennas. While standing, the torso height is also lowered.\\
     & X & Cancel all active animations. \\
     & Y & Turn on the background animation layer. \\
    Left Trackpad & & Trigger an episodic motion. Each quadrant of the trackpad maps to a different motion. \\
    Right Trackpad & & Like the left trackpad. Reserved to trigger four additional episodic motions. \\
    Backside & L1 & Turn the head lamp on and off. \\
    & R1 & Single press: Start and stop walking, Hold: increase the walking velocity gain to \SI{100}{\percent}. Without holding R1, all velocity commands are scaled to \SI{50}{\percent} of the maximum.\\
    & L2\,/\,R2 & During walking: Lateral walking velocity. During standing: Roll the torso while the head remains stationary. \\
    & L4 & Short press: trigger a happy animation. Long press: trigger an angry animation. \\
    & L5 & Short press: trigger an anxious animation. Long press: trigger a curious animation.\\
    & R4 & Short press: trigger a ``yes" animation. Long press: trigger a ``no" animation. \\
    & R5 & Trigger an expressive audio clip. \\\bottomrule
    \end{tabular}
\end{center}
\end{table}

\end{document}